\documentclass[pdflatex,sn-mathphys-ay]{sn-jnl}

\usepackage{graphicx}%
\usepackage{multirow}%
\usepackage{amsmath,amssymb,amsfonts}%
\usepackage{amsthm}%
\usepackage{mathrsfs}%
\usepackage[title]{appendix}%
\usepackage[dvipsnames]{xcolor}
\usepackage{textcomp}%
\usepackage{manyfoot}%
\usepackage{booktabs}%
\usepackage{algorithm}%
\usepackage{algorithmicx}%
\usepackage{algpseudocode}%
\usepackage{listings}%
\usepackage{enumitem}
\usepackage{subcaption}
\usepackage{tabularx}
\usepackage{pifont}

\newcommand{\up}[1]{{#1}}

\newcommand{\rev}[1]{{#1}}

\theoremstyle{thmstyleone}%
\theoremstyle{thmstyletwo}%

\theoremstyle{thmstylethree}%

\begin{document}

\title[Article Title]{Diffusion-Based Data Augmentation for Image Recognition: A Systematic Analysis and Evaluation}


\author[1]{\fnm{Zekun} \sur{Li}}\email{lizekun@smail.nju.edu.cn}

\author*[1]{\fnm{Yinghuan} \sur{Shi}}\email{syh@nju.edu.cn}

\author[1]{\fnm{Yang} \sur{Gao}}\email{gaoy@nju.edu.cn}

\author[2]{\fnm{Dong} \sur{Xu}}\email{dongxu@cs.hku.hk}

\affil[1]{\orgdiv{State Key Laboratory for Novel Software Technology}, \orgname{Nanjing University}, \orgaddress{\city{Nanjing}, \country{China}}}

\affil[2]{\orgdiv{School of Computing and Data Science}, \orgname{The University of Hong Kong}, \orgaddress{\city{Hong Kong SAR}, \country{China}}}


\abstract{Diffusion-based data augmentation (DiffDA) has emerged as a promising approach to improving classification performance under data scarcity. However, existing works vary significantly in task configurations, model choices, and experimental pipelines, making it difficult to fairly compare methods or assess their effectiveness across different scenarios. Moreover, there remains a lack of systematic understanding of the full DiffDA workflow. In this work, we introduce UniDiffDA, a unified analytical framework that decomposes DiffDA methods into three core components: model fine-tuning, sample generation, and sample utilization. This perspective enables us to identify key differences among existing methods and clarify the overall design space. Building on this framework, we develop a comprehensive and fair evaluation protocol, benchmarking representative DiffDA methods across diverse low-data classification tasks. Extensive experiments reveal the relative strengths and limitations of different DiffDA strategies and offer practical insights into method design and deployment. All methods are re-implemented within a unified codebase, with full release of code and configurations to ensure reproducibility and to facilitate future research.\footnote{Code repository at \url{https://github.com/nukezil/DiffDA-Eval}}}





\maketitle

\section{Introduction}\label{sec1:intro}

Data augmentation (DA) has long been recognized as a crucial technique for enhancing the generalization ability of machine learning models, especially when annotated training data is limited. The concept was initially explored in the context of early convolutional neural networks such as LeNet \citep{lecun1998lenet}, where simple geometric transformations like translations and rotations were applied to improve performance on handwritten digit classification. The impact of DA became particularly evident with the success of AlexNet \citep{krizhevsky2012alexnet} on the ImageNet dataset, where random cropping and horizontal flipping were key components of the training pipeline. Since then, a wide range of augmentation strategies have been proposed, ranging from heuristic methods like Cutout \citep{devries2017cutout}, Mixup \citep{zhang2018mixup}, and CutMix \citep{yun2019cutmix} to policy-driven approaches such as AutoAugment \citep{cubuk2019autoaugment} and RandAugment \citep{cubuk2020randaugment}, which use search algorithms to identify optimal augmentation schedules. A comprehensive overview of these developments can be found in \cite{shorten2019survey}.

While traditional DA techniques operate by transforming existing samples in input space, the use of generative models to synthesize novel data points presents a powerful alternative. Early works \citep{gurumurthy2017deligan, antoniou2018data, zhang2018imbalance} in this direction was primarily driven by generative adversarial networks (GANs) \citep{goodfellow2014gans}. Despite their promise, GAN-based augmentation methods suffer from fundamental limitations, including training instability, mode collapse, and limited semantic control \citep{arjovsky2017towards, metz2016unrolled, salimans2016improved}. These issues often result in synthetic images that lack sufficient diversity or fail to capture task-relevant fine-grained semantics \citep{odena2017conditional, isola2017image}.

More recent advances in generative modeling, particularly diffusion models, provide a new methodological pathway that alleviates many of the limitations associated with GANs \citep{yang2023diffusion}. Denoising Diffusion Probabilistic Models (DDPM) \citep{ho2020ddpm} represent a seminal contribution, introducing a simple yet stable training paradigm. Subsequent techniques such as Denoising Diffusion Implicit Models (DDIM) \citep{song2021ddim} and classifier-free guidance (CFG) \citep{ho2022classifierfree} have significantly improved sampling efficiency and controllability. Building upon these foundations, Latent Diffusion Models (LDMs) \citep{rombach2022stable} introduced a key innovation by conducting the diffusion process in a compact latent space, which greatly reduces computational demands while maintaining high-quality image generation. This innovation enabled the creation of Stable Diffusion, a series of widely adopted text-to-image models trained on large-scale image-text datasets. 

The strong generative capabilities and rich semantic priors embedded in Stable Diffusion have spurred the emergence of Diffusion-based Data Augmentation (DiffDA) methods \citep{he2022realguidance, zhang2023gif, islam2024diffusemix, trabucco2024dafusion, huang2024active, zhu2024distribution, kim2024datadream, wang2024diffmix, wang2025diffii}. These approaches leverage pre-trained diffusion models to synthesize diverse and semantically meaningful training samples, primarily aiming to address the challenge of limited data availability in image recognition. By harnessing the controllability and flexibility of diffusion-based synthesis, DiffDA offers a promising alternative to traditional augmentation and GAN-based methods, enabling more effective enhancement of downstream classification performance across various scenarios.

Despite the rapid progress in DiffDA, the current landscape remains fragmented and difficult to interpret. Prior works are typically evaluated under incompatible experimental setups that differ in key aspects such as dataset selection and split protocols, the choice of generative backbones, classifier architectures, and training strategies. These inconsistencies make it hard to draw reliable conclusions or to directly compare reported results across papers. Moreover, existing studies usually treat sample utilization as a minor implementation detail: some methods simply augment the training set by concatenating generated images with real data \citep{he2022realguidance, zhang2023gif}, while others randomly replace real samples with synthetic ones at a fixed probability during training \citep{trabucco2024dafusion,wang2024diffmix,wang2025diffii}. Systematic analyses of how these utilization strategies influence the effectiveness and efficiency of DiffDA remain lacking. \rev{Moreover, it is still unclear under what conditions DiffDA is truly beneficial, and whether existing approaches can generalize across varying data regimes, semantic granularities, and domain shifts.}
 
To address these issues, we introduce UniDiffDA, a unified framework for systematic evaluation and analysis of diffusion-based data augmentation. At the core of UniDiffDA is a modular decomposition of each DiffDA method into three components: diffusion model fine-tuning, sample generation, and sample utilization. This decomposition enables us to clearly identify and compare the technical choices made by different methods, and to reason about their interactions. \rev{Importantly, the framework is not limited to summarizing existing practices: it also guides us to design general techniques that improve DiffDA performance and efficiency along all three components. We conduct comprehensive experiments across a broad spectrum of low-data image recognition tasks covering different semantic granularities and domains, ranging from generic object classification and fine-grained recognition to long-tailed and multi-domain benchmarks.}

\rev{Building on this framework, we conduct an extensive empirical study to characterize the strengths and limitations of representative DiffDA methods across diverse tasks. We systematically investigate how sample utilization strategies, critical hyperparameters, and generative backbones influence the performance of DiffDA, and we derive actionable insights for future research and practical deployment. Beyond benchmarking, we further explore general techniques from the perspectives of model fine-tuning, sample generation, and sample utilization, demonstrating that appropriately designed prompts, accelerated sampling schemes, and simple yet effective filtering rules can, under specific scenarios, make existing DiffDA methods both stronger and faster.}

We summarize the core contributions of this work as follows:
\begin{itemize}[topsep=2pt, itemsep=2pt]
\item We introduce a \textbf{unified analytical perspective} on DiffDA, highlighting the differences among existing methods in terms of fine-tuning requirements, generation strategies, and data utilization techniques.
\item We establish a \textbf{comprehensive and fair evaluation protocol}, supported by extensive experiments that benchmark representative DiffDA methods across diverse low-data scenarios.
\item \rev{We explore \textbf{general methodological techniques} along all three DiffDA components to improve both the effectiveness and efficiency of existing methods.}
\item We release a \textbf{fully open-sourced and reproducible codebase}, providing a reproducible benchmark to facilitate future research and applications in DiffDA.
\end{itemize}

The remainder of this paper is organized as follows. Section \ref{sec2:background} reviews the background on diffusion models relevant to diffusion-based data augmentation. Section \ref{sec3:methods} details the proposed UniDiffDA framework and explains how representative methods are grouped and examined under this unified view, together with the evaluation protocol employed in our experiments. Section \ref{sec4:exp} presents experimental results, followed by comprehensive analysis and discussion. Finally, Section \ref{sec5:conclusion} summarizes the paper.

\section{Background}\label{sec2:background}
\subsection{Diffusion-Based Text-to-Image Generation}
Diffusion models represent a class of generative models that synthesize data by learning to reverse a gradual corruption process. The conceptual foundation of these models traces back to nonequilibrium thermodynamics \citep{sohl2015deep}, where data generation is modeled as a Markov chain that gradually removes noise from an initial random sample. This approach was later formalized through score-based generative modeling \citep{song2019generative} and refined in the Denoising Diffusion Probabilistic Model (DDPM) framework \citep{ho2020ddpm}.
In DDPM, a forward process gradually adds Gaussian noise to a clean data sample $\mathbf{x}_0$ over $T$ steps:
\begin{equation}
    q(\mathbf{x}_t|\mathbf{x}_{t-1})=\mathcal{N}(\mathbf{x}_t;\sqrt{1-\beta_t}\mathbf{x}_{t-1},\beta_t\mathbf{I}),\quad t=1,\ldots,T, 
\end{equation}
where $\beta_t$ denotes the variance schedule. The overall marginal distribution for $\mathbf{x}_t$  given the original image is:
\begin{equation}
\label{eq:diffuse}
    q(\mathbf{x}_t|\mathbf{x}_0)=\mathcal{N}(\mathbf{x}_t;\sqrt{\bar{\alpha}_t}\mathbf{x}_0,(1-\bar{\alpha}_t)\mathbf{I}),\quad\mathrm{with~}\bar{\alpha}_t=\prod_{i=1}^t(1-\beta_i).
\end{equation}
The denoising process is learned by training a neural network $\boldsymbol{\epsilon}_\theta$ to predict the noise added at each step using a simple mean squared error loss:
\begin{equation}
    \mathcal{L}_{\mathrm{DDPM}}=\mathbb{E}_{\mathbf{x}_0,\boldsymbol{\epsilon}\sim\mathcal{N}(\mathbf{0},\mathbf{I}),t}\left[\left\|\boldsymbol{\epsilon}-\boldsymbol{\epsilon}_\theta(\sqrt{\bar{\alpha}_t}\mathbf{x}_0+\sqrt{1-\bar{\alpha}_t}\boldsymbol{\epsilon},t)\right\|^2\right].
\end{equation}
Once trained, the model samples from the data distribution by starting from Gaussian noise $\mathbf{x}_T\sim\mathcal{N}(\mathbf{0},\mathbf{I})$ and applying the learned reverse process:
\begin{equation}
\label{eq:denoise}
    \mathbf{x}_{t-1}=\frac{1}{\sqrt{\alpha_t}}\left(\mathbf{x}_t-\frac{1-\alpha_t}{\sqrt{1-\bar{\alpha}_t}}\boldsymbol{\epsilon}_\theta(\mathbf{x}_t,t)\right)+\sigma_t\mathbf{z},\quad\mathbf{z}\sim\mathcal{N}(\mathbf{0},\mathbf{I}),
\end{equation}
where $\sigma_t$ controls the level of added stochasticity at each step.

While DDPM achieves high generation quality, it requires hundreds to thousands of reverse steps for inference. To reduce the high computational cost, subsequent works such as DDIM \citep{song2021ddim}, DPM-Solver \citep{lu2022dpm_solver}, and DPM-Solver++ \citep{lu2022dpm_solver_pp} introduce deterministic and accelerated solvers that retain sample quality with much fewer steps. In parallel, classifier-free guidance (CFG) \citep{ho2022classifierfree} improves conditional sample fidelity without requiring an external classifier, enabling effective control via text prompts.

Building on these technical foundations, the Latent Diffusion Model (LDM) \citep{rombach2022stable} proposes conducting the denoising process in a compressed latent space. Specifically, a pretrained variational autoencoder (VAE) is used to encode an input image $\mathbf{x}$ into a latent vector $\mathbf{z}=\mathcal{E}(\mathbf{x})$, upon which a UNet-based denoising network is trained. The training objective becomes:
\begin{equation}
    \mathcal{L}_{\mathrm{LDM}}=\mathbb{E}_{\mathbf{x},t,\boldsymbol{\epsilon}}\left[\left\|\boldsymbol{\epsilon}-\boldsymbol{\epsilon}_\theta(\mathbf{z}_t,t,\tau)\right\|^2\right],
\end{equation}
where $\mathbf{z}_t$ is the noisy latent vector and $\tau$ denotes an optional conditioning signal, such as CLIP-encoded text prompts. LDM enables efficient high-resolution image generation by operating in a learned latent space, and provide the architectural foundation for Stable Diffusion. The Stable Diffusion model combines a frozen CLIP text encoder, a VAE encoder-decoder, and a text-conditioned UNet denoiser. Pre-trained on hundreds of millions of image-text pairs, it offers strong semantic control and generation quality, and has been widely adopted in diverse downstream tasks. In this paper, we primarily adopt Stable-Diffusion-v1.5\footnote{\url{https://huggingface.co/stable-diffusion-v1-5/stable-diffusion-v1-5}} as the foundation generative model for all DiffDA methods, consistent with most recent studies \citep{wang2024diffmix, wang2025diffii}, \rev{and additionally investigate the impact of replacing it with more advanced backbones such as Stable-Diffusion-2.1\footnote{\url{https://huggingface.co/stabilityai/stable-diffusion-2-1}} and Stable-Diffusion-3.5-medium\footnote{\url{https://huggingface.co/stabilityai/stable-diffusion-3.5-medium}}.}

\subsection{Diffusion-Based Image-to-Image Transition}
While text-to-image diffusion models can generate diverse and semantically rich samples, directly using them for data augmentation often leads to suboptimal results due to distribution mismatch with real training data. Generated images may lack alignment with the original data domain or fail to preserve task-relevant semantics. To mitigate this issue, an image-to-image transition paradigm has become the mainstream choice in current DiffDA methods. Instead of generating samples from scratch, it applies diffusion models to transform real images into augmented variants. This strategy helps preserve semantic consistency with the original data while introducing meaningful diversity, making it better aligned with the goals of data augmentation.

One of the most widely adopted techniques for diffusion-based image-to-image transition is SDEdit \citep{meng2022sdedit}. The key idea is to start denoising from a partially noised version of the input image, rather than from pure Gaussian noise. Formally, given a clean image $\mathbf{x}_{0}$, a strength parameter $s\in(0,1]$ controls how many steps of the forward diffusion process are applied to obtain the noisy image $\mathbf{x}_{sT}$, where $T$ is the total number of diffusion steps. This corresponds to computing $\mathbf{x}_{sT}\sim q(\mathbf{x}_t|\mathbf{x}_0)$ with $t=sT$, following the forward process defined in Eq.~\ref{eq:diffuse}. The partially noised image $\mathbf{x}_{sT}$ then serves as the starting point for the reverse sampling procedure described in Eq.~\ref{eq:denoise}, which generate an output image $\mathbf{x}_0^{\prime}$. By tuning the value of $s$, one can control the degree of transformation: lower $s$ values yield outputs closer to the original input, while higher $s$ values introduce greater variability and semantic changes.

\begin{figure}[htbp]
\centering
\includegraphics[width=\textwidth]{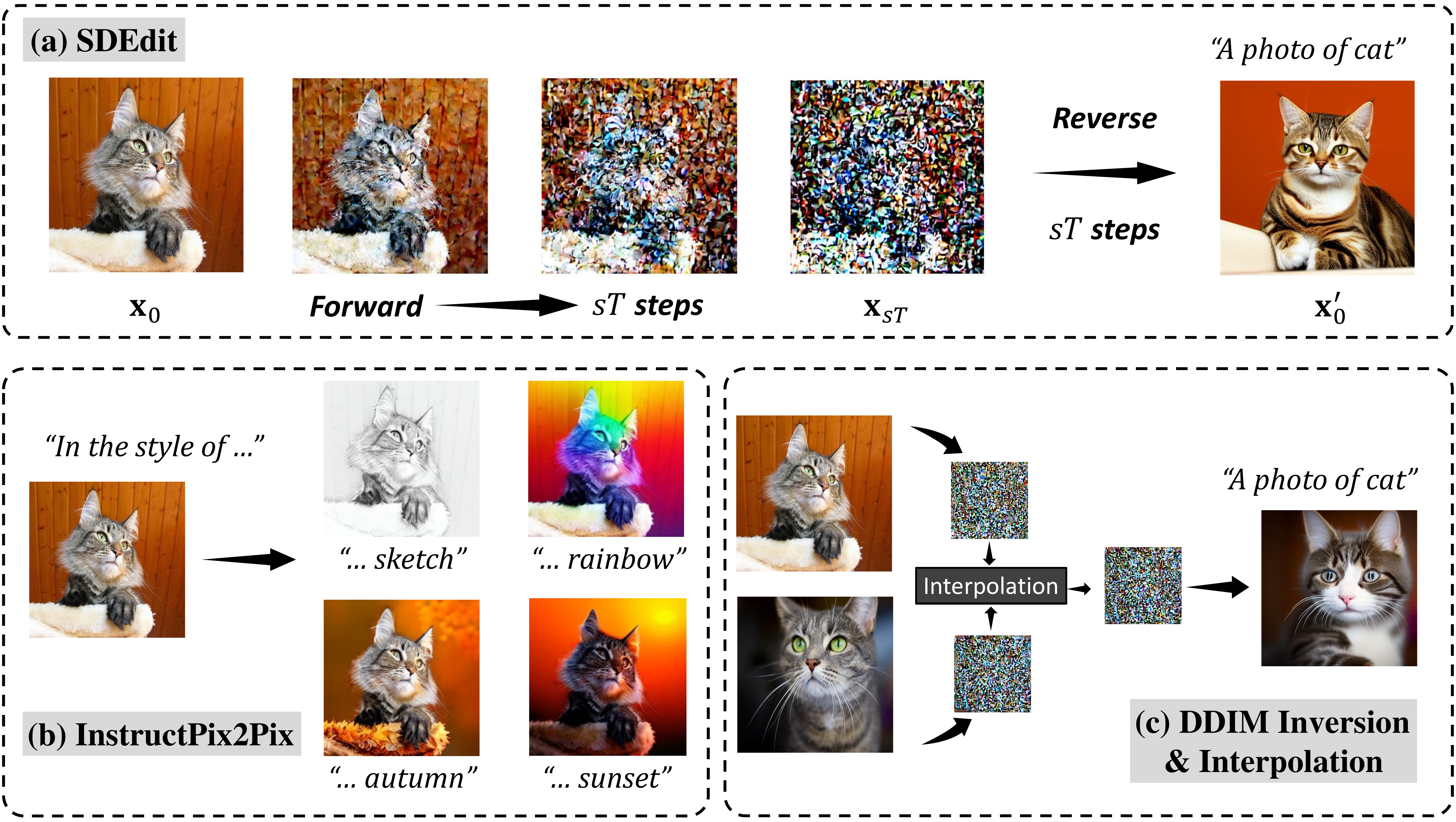}
\caption{Illustration of three diffusion-based image-to-image transition techniques used in representative DiffDA methods: (a) SDEdit, (b) InstructPix2Pix, and (c) DDIM Inversion and Interpolation.}\label{fig:img2img}
\end{figure}

Beyond SDEdit, some DiffDA methods adopt alternative image-to-image transition strategies. InstructPix2Pix \citep{brooks2023instructpix2pix} fine-tunes the Stable Diffusion model using paired images and textual instructions, allowing it to respond to editing prompts such as ``\texttt{add a giant red dragon}'' or ``\texttt{in the style of a coloring book}''. This enables controllable augmentation based on natural language. DDIM inversion \citep{song2021ddim} deterministically maps an image back to its latent noise representation through a reverse DDIM sampling process. Unlike SDEdit, which adds stochastic noise to an image before denoising, DDIM inversion provides a consistent latent code that can be edited and re-synthesized, facilitating latent space interpolation for label-preserving generation. We provide illustrative comparisons of these three techniques in Fig.~\ref{fig:img2img}. We will elaborate in later sections on how these image-to-image transition techniques are concretely applied to data augmentation in various DiffDA methods.

\subsection{Fine-tuning Diffusion Models}
Although the pretrained Stable Diffusion model has acquired strong generative priors from large-scale image-text datasets, enabling them to synthesize diverse and realistic images for many common concepts such as ``\texttt{airplane}'', ``\texttt{bird}'', or ``\texttt{cat}'', its capability becomes limited when dealing with fine-grained categories (e.g., ``\texttt{Sage Thrasher}'', a specific bird species) or domain-specific concepts (e.g., ``\texttt{lymphocyte}'', a type of white blood cell). In such cases, the model often fails to generate semantically meaningful or visually coherent images that can serve as effective training data for downstream classifiers. In such cases, the model often fails to generate semantically meaningful or visually coherent images that can serve as effective training data for downstream classifiers. This limitation motivates the need to fine-tune diffusion models using real examples from the target domain, allowing them to better capture class-specific attributes and produce more relevant synthetic data for augmentation.

\begin{figure}[htbp]
\centering
\includegraphics[width=\textwidth]{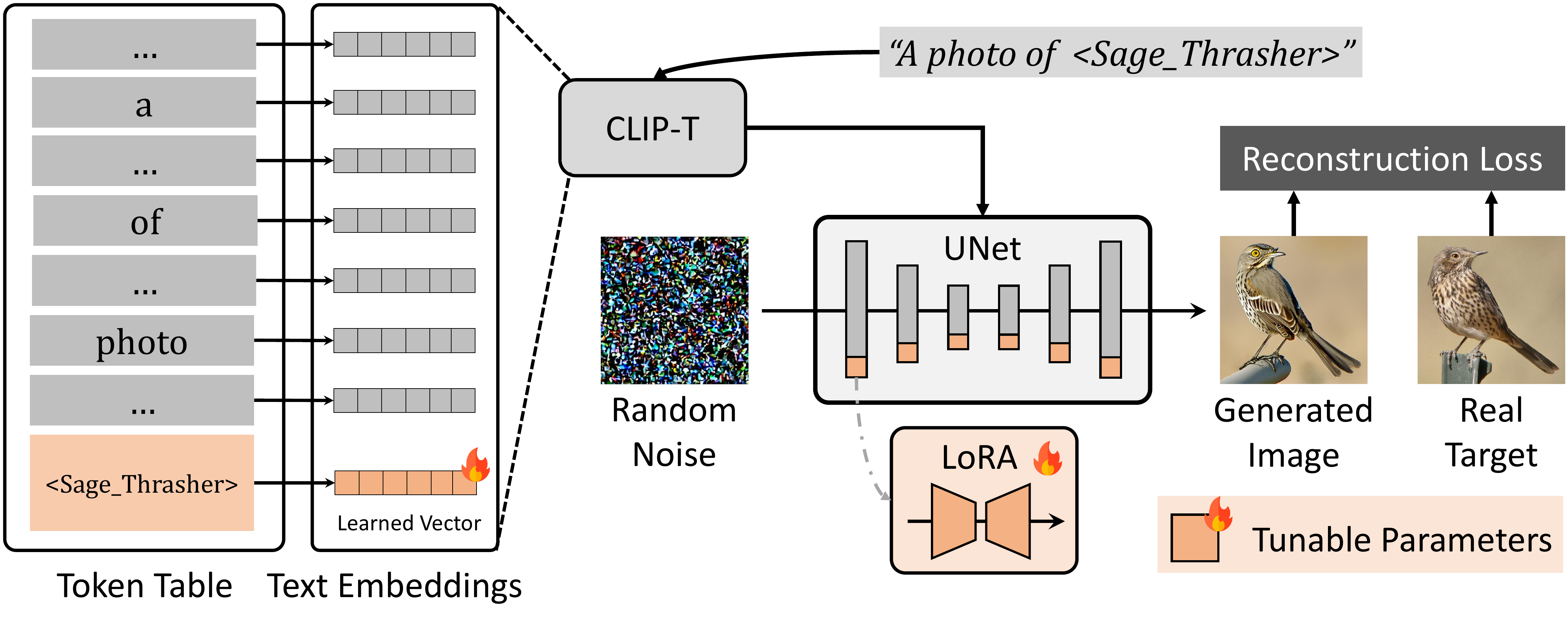}
\caption{Illustration of the fine-tuning pipeline combining Textual Inversion and DreamBooth-LoRA. A pseudo-token (e.g., \texttt{<Sage\_Thrasher>}) is learned and inserted into the text prompt to synthesize class-specific images. The corresponding embedding is optimized while keeping the rest of the text encoder frozen. Meanwhile, LoRA modules are inserted into the UNet and fine-tuned to better capture target-domain visual features. The generated image is trained to match the real target image via reconstruction loss. Tunable parameters are highlighted in orange.}\label{fig:finetune}
\vspace{-0.3cm}
\end{figure}

A common technique for fine-tuning diffusion models with limited target data is Textual Inversion \citep{gal2023image}. It enables the model to synthesize class-specific images by learning a new pseudo-token, typically denoted as ``\texttt{<class-name>}'',  which is inserted into the text prompt (e.g., ``\texttt{a photo of <class-name>}''). This token is associated with a trainable embedding vector, while the rest of the model remains frozen. The embedding is optimized using a few example images from the target class, so that conditioning on the prompt containing ``\texttt{<class-name>}'' results in images capturing the desired visual concept. While Textual Inversion preserves the original denoising network and only learns a lightweight embedding, its generation fidelity can be limited when modeling visually complex or fine-grained concepts. A natural extension is to also fine-tune the denoising UNet of Stable Diffusion. This approach is commonly referred to as DreamBooth \citep{ruiz2023dreambooth} in existing DiffDA literature\footnote{The way DreamBooth is applied in DiffDA is not exactly the same as the original formulation proposed by \cite{ruiz2023dreambooth}, which focuses on subject-driven image personalization. We just follow the established terminology and use DreamBooth to denote the fine-tuning of the UNet for novel category synthesis.} \citep{wang2024diffmix,wang2025diffii}. However, fully fine-tuning the UNet in Stable Diffusion is often computationally expensive. To reduce overhead, recent DiffDA methods commonly adopt Low-Rank Adaptation (LoRA) \citep{hu2022lora}, which inserts small trainable low-rank matrices into the UNet while keeping most weights frozen. We provide an illustration of the aforementioned fine-tuning strategies in Fig.~\ref{fig:finetune}.

\section{Diffusion-Based Data Augmentation}\label{sec3:methods}
In this section, to better understand the design space of DiffDA methods, we first propose a unified analytical framework that dissects each method into three modular components. This decomposition allows for a structured comparison and lays the foundation for our systematic evaluation. We then introduce several representative methods under this framework and conclude the section with the evaluation protocol adopted throughout our study.

\begin{figure}[htbp]
\centering
\includegraphics[width=\textwidth]{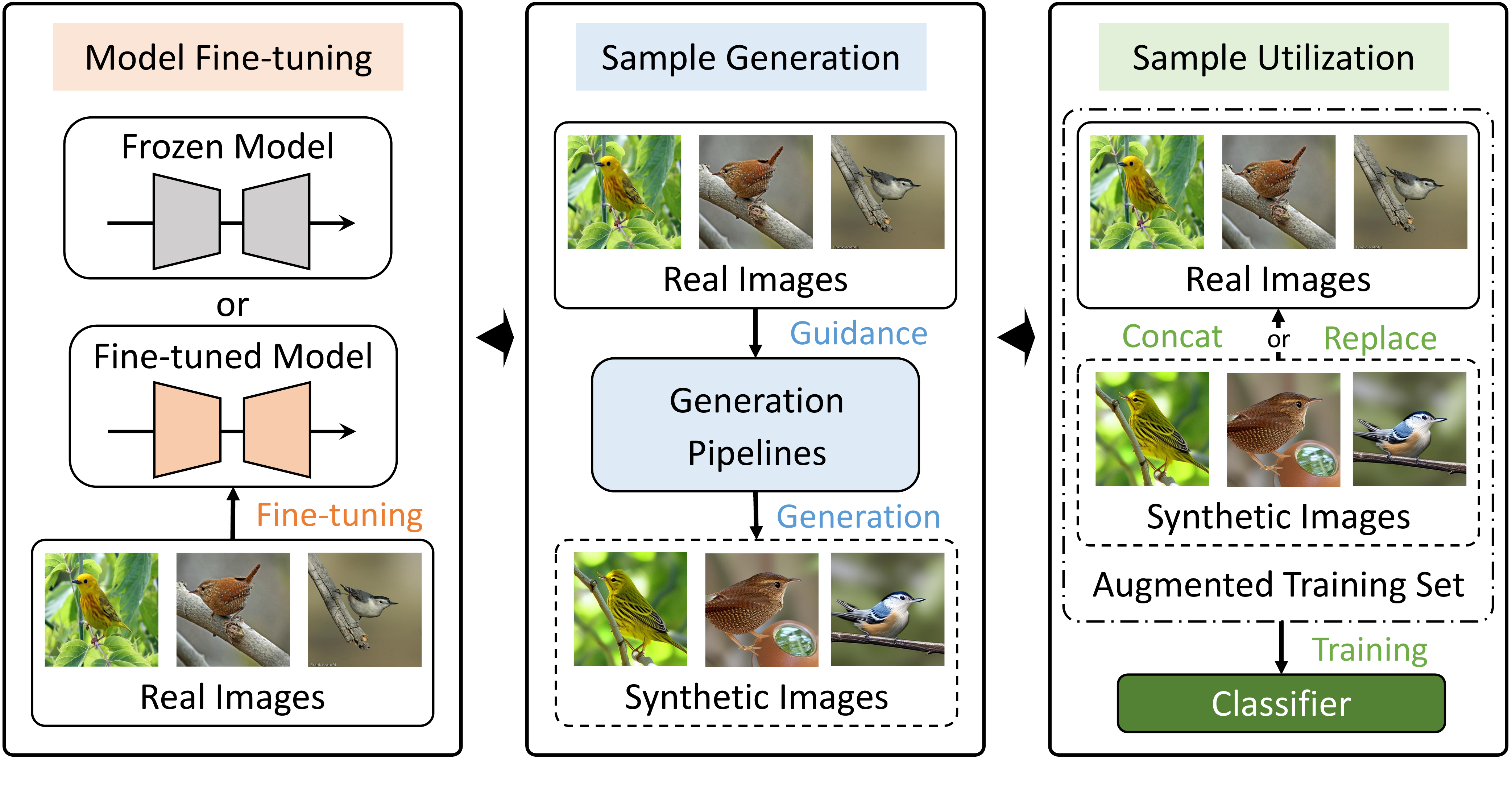}
\caption{Illustration of the UniDiffDA framework, which decomposes diffusion-based data augmentation (DiffDA) workflow into three core components: (1) \textbf{Model Fine-tuning}, where the diffusion model is optionally adapted to the target domain using real images; (2) \textbf{Sample Generation}, where the model synthesizes new samples guided by real data; and (3) \textbf{Sample Utilization}, where synthetic samples are either concatenated with or used to replace real samples for classifier training.}\label{fig:framework}
\end{figure}

\subsection{The UniDiffDA Framework} 
While DiffDA has gained traction across a variety of low-data image classification tasks, existing methods are often designed independently, making it difficult to conduct systematic analysis or fair comparisons. To address this gap, we propose the UniDiffDA framework, a modular perspective that decomposes any DiffDA method into three essential and sequential components: (1) diffusion model fine-tuning, (2) sample generation, and (3) sample utilization, as illustrated in Fig.~\ref{fig:framework}. This unified view not only provides a structured way to understand existing techniques but also enables principled comparisons and design choices across different settings.

\textbf{Fine-tuning the diffusion model.} 
An important design decision in diffusion-based data augmentation is whether and how to adapt the generative model to the target dataset. Fine-tuning can help align the model’s prior with the specific semantics of the downstream task, improving the relevance and fidelity of generated samples. However, this is not always advantageous. For well-represented categories already covered in pretraining (e.g., ``airplanes'' or ``dogs''), fine-tuning on limited target data may lead to overfitting and reduce sample diversity. Additionally, model adaptation incurs extra training cost. More sophisticated adaptation techniques may amplify both the risk of overfitting and the associated computational overhead.

\textbf{Generating synthetic samples.}
The core design space in this stage lies in the choice of image-to-image transition strategy to generate synthetic samples. Among the most widely adopted approaches is SDEdit-style partial denoising, where a strength parameter $s$ controls how much noise is added before the reverse process begins. This parameter governs the degree of deviation from the original image, balancing realism and augmentation diversity. Another important design aspect is the formulation of text prompts used to condition the generation process. Simple prompts typically follow a fixed template that includes only the class name (e.g., ``\texttt{a photo of cat}''), while more complex prompts may involve handcrafted descriptions or detailed sentences generated by large language models. Some methods introduce customizations to the diffusion process itself, such as modifying the initialization of the noise input. Others apply specific post-processing techniques to the generated images to improve their visual diversity. We defer the detailed explanation of such strategies to the next subsection, where each representative method is introduced individually.

\textbf{Utilizing synthetic samples.}
Once synthetic samples are generated, a key design choice lies in how to incorporate them into the classifier training. Let the original labeled dataset be denoted as $\mathcal{X} = {(\mathbf{x}_i, y_i)}_{i=1}^N$, and let the diffusion model generate $M$ synthetic variants $\{(x_{i1}^{\prime},y_{i1}^{\prime}),(x_{i2}^{\prime},y_{i2}^{\prime}),\ldots,(x_{iM}^{\prime},y_{iM}^{\prime})\}$ for each $\mathbf{x}_i$. In most cases, the generated labels are inherited from the originals, i.e., $y_{ij}^{\prime}=y_i$, assuming label preservation during synthesis, while some strategies may assign modified labels in a Mixup \citep{zhang2018mixup} style. We summarize four typical strategies for utilizing $\mathcal{X}' = \bigcup_{i=1}^N \{(\mathbf{x}_{ij}', y_{ij}')\}_{j=1}^M$ during classifier training:
\begin{itemize}[topsep=2pt, itemsep=2pt]
    \item \textbf{Full Concatenation}. The simplest approach is to directly merge the original and generated data: $\mathcal{X}_{\mathrm{train}}=\mathcal{X}\cup\mathcal{X}^{\prime}$. This allows full retention of accurate supervision from real data and complete utilization of synthetic data. However, it increases the training set size by a factor of by a factor of $1{+}M$, leading to significantly higher training time and resource consumption.
    \item \textbf{Full Replacement}. In this case, the original dataset is completely replaced by the generated one: $\mathcal{X}_{\mathrm{train}}=\mathcal{X}^{\prime}$. This reduces training time compared to concatenation. However, it discards the original samples entirely, which may result in performance drop if the synthetic samples contain semantic noise or artifacts.
    \item \textbf{Local Random Replacement}. For each original sample $(\mathbf{x}_i, y_i)$, one of its $M$ variants $(\mathbf{x}_{ij}', y_{ij}')$ may replace it with a fixed probability $p$ in each training epoch:
    \begin{equation}
        (x_i^{\mathrm{train}},y_i^{\mathrm{train}})=\begin{cases}(\mathbf{x}_{ij}', y_{ij}'),&\text{with probability }p,\\
        (\mathbf{x}_i, y_i),&\text{with probability }1-p.
        \end{cases}
    \end{equation}
    This strategy maintains the original dataset size, thereby keeping training cost close to baseline. However, it can introduce supervision noise when inaccurate synthetic samples are used in place of clean real ones.
    \item \textbf{Global Random Replacement}. Instead of choosing replacements from each sample’s own augmented variants, synthetic samples are drawn from the global pool $\mathcal{X}'$:
    \begin{equation}
        (x_i^{\mathrm{train}},y_i^{\mathrm{train}})=\begin{cases}(\mathbf{x}_{ij}', y_{ij}'),&\text{with probability }p, (\mathbf{x}_{ij}', y_{ij}') \sim \mathcal{X}'\\
        (\mathbf{x}_i, y_i),&\text{with probability }1-p.
        \end{cases}
    \end{equation}
    Compared to Local Random Replacement, this strategy is easier to implement, as it does not require maintaining correspondence between real and synthetic samples. It may also introduce more stochasticity into training.
\end{itemize}
In summary, each sample utilization strategy reflects a trade-off between training efficiency and augmentation effectiveness. While critical to the overall performance of DiffDA, this component remains underexplored in current research.

\begin{table}[htbp]
\centering
\caption{Summary of representative DiffDA methods and their key technical designs under the UniDiffDA framework.}\label{tab:methods}
{\setlength{\tabcolsep}{3pt}
\begin{tabularx}{\textwidth}{@{}llX@{}}
\toprule
\textbf{Method} & \textbf{Venue} & \textbf{Key Technical Designs} \\
\midrule

 &  & \ding{59} No fine-tuning \\
Real Guidance \citep{he2022realguidance} & ICLR 2023 & \ding{70} SDEdit, single simple prompt \\
& & \ding{86} Full Concatenation \\
\midrule
\addlinespace[0.5ex]

 &  & \ding{59} No fine-tuning \\
GIF \citep{zhang2023gif} & NeurIPS 2023 & \ding{70} SDEdit, prompt pooling, noise optimization \\
& & \ding{86} Full Concatenation \\
\midrule
\addlinespace[0.5ex]

&  &
\ding{59} No fine-tuning  \\
DiffuseMix \citep{islam2024diffusemix} & CVPR 2024 & \ding{70} InstructPix2Pix, prompt pooling\footnotemark[1] \\
& & \ding{86} Full Replacement \\
\midrule
\addlinespace[0.5ex]

 &  & \ding{59} Textual Inversion \\
DA-Fusion \citep{trabucco2024dafusion} & ICLR 2024 & \ding{70} SDEdit, single simple prompt \\
& & \ding{86} Local Random Replacement \\
\midrule
\addlinespace[0.5ex]

 &  &
\ding{59} Textual Inversion + DreamBooth-LoRA  \\
Diff-Aug \citep{wang2024diffmix} & CVPR 2024 & \ding{70} SDEdit, single simple prompt \\
& & \ding{86} Global Random Replacement \\
\midrule
\addlinespace[0.5ex]

 &  &
\ding{59} Textual Inversion + DreamBooth-LoRA \\
\multirow{2}{*}{Diff-Mix \citep{wang2024diffmix}} & \multirow{2}{*}{CVPR 2024} & \ding{70} SDEdit, single simple prompt \\
& &  \ding{70} Inter-class image mixup \\
& & \ding{86} Global Random Replacement \\
\midrule
\addlinespace[0.5ex]

 &  &
\ding{59} Textual Inversion + DreamBooth-LoRA \\
\multirow{2}{*}{Diff-II \citep{wang2025diffii}} & \multirow{2}{*}{CVPR 2025} & \ding{70} DDIM inversion and latent interpolation \\
& &  \ding{70} LLM-enhanced prompt pooling \\
& & \ding{86} Local Random Replacement \\

\bottomrule
\end{tabularx}
}
\footnotetext{\parbox{\linewidth}{
    \raggedright
    \footnotesize
    We use \ding{59}, \ding{70}, and \ding{86} to denote technical details related to model fine-tuning, sample generation, and sample utilization, respectively.\\
    \textsuperscript{1}\hspace{0.1em}Prompt pooling refers to randomly selecting a prompt from a curated prompt set.\\
  }}
\end{table}

\subsection{Representative Methods}
We now introduce a set of representative DiffDA methods, categorized under the UniDiffDA framework. These methods illustrate diverse design choices across model fine-tuning, synthetic data generation, and sample utilization. An overview of their key technical designs is summarized in Table~\ref{tab:methods}.

\textbf{Real Guidance} \citep{he2022realguidance} is one of the earliest methods explore diffusion-based data augmentation. It does not perform any fine-tuning on the diffusion model. For sample generation, it adopts the SDEdit-style strategy, where each real image is noised for $sT$ steps and then denoised using a class-specific prompt. The prompt is a simple template such as ``\texttt{a photo of cat}'', containing only the class name. For sample utilization, it applies Full Concatenation, augmenting the training set by directly appending synthetic samples to real data.

\textbf{GIF} (Guided Imagination Framework) was introduced by \cite{zhang2023gif}, which also follows SDEdit but uniquely optimizes the noisy latent initialization used in the denoising process. This optimization aims to improve class-maintained informativeness, which is evaluated using an auxiliary CLIP \citep{radford2021clip} model, and sample diversity, measured by the KL divergence between generated and original samples. For text prompts, GIF carefully constructs them by combining random domain labels and adjective words, for example, ``\texttt{an oil painting of colorful dog}'', where ``\texttt{an oil painting of}'' and ``\texttt{colorful}'' come from curated lists. The method uses the pretrained diffusion model without adaptation and also adopts the Full Concatenation strategy for sample utilization.

\textbf{DiffuseMix} \citep{islam2024diffusemix} adopts a more structured image mixing approach driven by a pretrained diffusion model without fine-tuning. It employs a label-preserving image-to-image augmentation strategy based on InstructPix2Pix. For each image, DiffuseMix applies a text-guided transformation using prompts that describe style changes without referring to class labels, such as ``\texttt{a transformed version of image into autumn}'' or ``\texttt{...into watercolor art}''. A hybrid image is constructed by concatenating a portion of the original image with the complementary portion from the generated image. This hybrid is then blended with a randomly selected fractal image, producing the final training image with enhanced structural diversity. DiffuseMix adopts Full Replacement for sample utilization, based on the assumption that each generated image is expected to preserve part of the original supervision.

\textbf{DA-Fusion} was proposed by \cite{trabucco2024dafusion}, aiming to address the limited ability of pretrained diffusion models to synthesize novel visual concepts. It adopts Textual Inversion to optimize one pseudo-token for each target-domain category. In data generation, it follows the standard SDEdit paradigm. Text prompts are structured as ``\texttt{a photo of a <class-name>}'', where ``\texttt{<class-name>}'' refers to the learned pseudo-token. DA-Fusion employs Local Random Replacement for sample utilization.

\textbf{Diff-Mix} \citep{wang2024diffmix} leverages a fine-tuned diffusion model where both the text embeddings and UNet are adapted using Textual Inversion and DreamBooth-LoRA. For sample generation, it performs inter-class image mixup with SDEdit: given an image from class A, it uses a prompt describing class B to guide the diffusion model to synthesize a new image that retains the background of A while replacing the foreground with B. For fine-grained datasets, Diff-Mix constructs prompts by combining class-specific pseudo-tokens with their metaclass, such as ``\texttt{a photo of a <American Crow> bird}''. For sample utilization, Diff-Mix adopts Global Random Replacement. \cite{wang2024diffmix} also propose a baseline method named \textbf{Diff-Aug}, which uses the fine-tuned diffusion model to generate label-preserving synthetic samples.

\textbf{Diff-II} \citep{wang2025diffii} is one of the most recent DiffDA methods and adopts the same fine-tuning strategy as Diff-Mix. For sample generation, it first applies DDIM inversion to map real images into the latent space, then selects two same-class samples and performs spherical interpolation between their inverted latents. A two-stage denoising process with a total of $T$ steps is used: the first stage runs for $(1-r)T$ steps and is guided by a suffixed prompt like ``\texttt{a photo of a <Winter Wren> bird standing on a tree branch}'', where the suffix is a detailed description generated by a large language model (LLM); the second stage removes the suffix and continues for the remaining $rT$ steps to refine the output. For sample utilization, Diff-II adopts Local Random Replacement, following DA-Fusion.

\subsection{Evaluation Protocol}
We present the key components of our systematic and fair evaluation of DiffDA methods, including comprehensive dataset selection, unified choices of generative models and classifier backbones, and consistent principles for method implementation.

\subsubsection{Datasets} 
We evaluate DiffDA methods across five natural image datasets and two medical image datasets. The natural image datasets include \textbf{Caltech-101} \citep{fei2007learning}, \textbf{CIFAR-100} \citep{krizhevsky2009learning}, \textbf{ImageNet100} and \rev{\textbf{ImageNet-1K}} \citep{deng2009imagenet}, representing coarse-grained classification tasks at different scales. In addition, we incorporate two fine-grained benchmarks: CUB-200-2011 (\textbf{Birds}) \citep{wah2011cub} and FGVC-Aircraft (\textbf{Aircraft}) \citep{maji2013fgvc_aircraft}. We further evaluate on two medical image datasets: the \textbf{Blood} dataset, comprising microscopic images of 8 types of peripheral blood cells \citep{acevedo2019recognition}, and the \textbf{Skin} dataset, consisting of dermatoscopic images of pigmented skin lesions across 7 categories \citep{tschandl2018ham10000}. \rev{Finally, to cover more challenging scenarios, we include the \textbf{Semi-iNat} dataset \citep{su2021semi_iNat} with labels at different semantic granularities and the \textbf{DomainNet} dataset \citep{peng2019moment}, which spans multiple visual domains.}

To better evaluate the effectiveness of DiffDA methods, we simulate data-scarce settings by limiting the number of training samples per class across all datasets. The specific setup and task configurations for each dataset will be introduced in the subsequent experimental section.

\subsubsection{Models}
While some original DiffDA methods were built upon diffusion models like GLIDE \citep{nichol2022glide} or DALL-E 2 \citep{Ramesh2022DALLE2}, we adopt Stable Diffusion v1.5 as a unified generative foundation model across all methods to ensure fair comparison. Notably, the InstructPix2Pix model\footnote{\url{https://huggingface.co/timbrooks/instruct-pix2pix}} used in DiffuseMix is also trained on Stable Diffusion v1.5. For classification, we primarily use ResNet-50 \citep{he2016residual}. Whether the classifier is trained from scratch or initialized with ImageNet-pre-trained weights depends on the specific task setting. In addition to ResNet-50, we further evaluate the performance of DiffDA methods when using MobileNetV3 \citep{howard2019searching} and ViT-B/16 \citep{dosovitskiy2020image} as classifier backbones.

\subsubsection{Methods}
\rev{For each DiffDA method, we first perform a systematic search and analysis of sample utilization strategies and shared hyperparameters based on a set of representative methods, and then adopt the resulting best-performing configurations when reporting our final results. For every task, all methods are evaluated under strictly matched experimental settings, including dataset splits, training schedules, and shared hyperparameters, in order to ensure fair comparison. Method-specific hyperparameters that are unique to a particular approach are kept consistent with their original implementations.}

\rev{\subsection{Additional Related Works}
In addition to the representative methods that we have analyzed in detail, there exist several other studies exploring diffusion-based data augmentation for image classification. Although these methods are not included in our experimental evaluation because their experimental designs differ substantially from the mainstream framework considered in this paper, we believe discussing them here can offer complementary insights. Several recent studies introduce additional conditional controls to diffusion models beyond text prompts, leveraging the ControlNet \citep{Zhang2023_ControlNet} architecture. \citet{Ma2024_3D_DST} render multi-view 2D images from CAD models that provide explicit 3D shape and pose information, then extract their edge maps as conditional inputs to ControlNet. \citet{Michaeli2024_SaSPA} instead derive structural edges directly from real images. In both cases, ControlNet offers enhanced structure-preserving capability, which helps improve the fidelity of generated samples. Another line of work \citep{huang2024active, Ma2025_InstructModelFails, Li2025_GenDataAgent} focuses on leveraging feedback from the current classifier during training to identify mis-classified or “hard” real samples and then selectively augmenting only those samples. This classifier-informed augmentation paradigm explicitly strengthens the classifier’s ability on difficult cases. However, the iterative generation process introduces significantly higher computational cost owing to repeated rounds of generation and training. There are also recent studies \citep{zhu2024distribution, Yuan2024_RealFakeTrainingDataSynthesis} that examine, from a theoretical standpoint, how to generate synthetic data whose distribution closely matches that of real training data.
}

\rev{
Beyond classification, diffusion-based data augmentation has been applied to more complex vision tasks, such as segmentation \citep{yang2023freemask, Toker2024_SatSynth, zhao2025pseudo} and detection \citep{li2024simple, zhu2024odgen, vu2025multi}. These applications typically require additional conditional controls, such as mask guidance, bounding boxes or multi-object layout constraints, to ensure proper alignment between generated samples and task-specific ground-truth. Although these studies are outside the scope of this paper, exploring how generative augmentation methods can be unified across different vision tasks remains a meaningful direction for future research.
}

\section{Experiments}\label{sec4:exp}
In this section, we present comprehensive experiments to evaluate and analyze representative DiffDA methods. We first describe the experimental setup, followed by the main results on various benchmark tasks. We then conduct in-depth analyses of key factors such as hyperparameter choices and finally summarize actionable insights for future DiffDA design and deployment. 

\subsection{Experimental Setup}
We begin by introducing the tasks used to benchmark the performance of DiffDA methods. Shared experimental settings across methods and tasks are reported here, while method-specific or task-dependent configurations will be discussed alongside the corresponding results. Our codebase is released with detailed configuration files to facilitate reproducibility. Additional implementation details that are not described in the main text are available in the released code.

\textbf{Task Configurations}. We evaluate DiffDA methods on a diverse set of classification tasks spanning coarse-grained, fine-grained, and medical domains. All models are trained on low-data subsets of the original datasets, along with their corresponding generated samples. For coarse-grained classification, we construct subsets of Caltech-101, CIFAR-100, ImageNet-100, and \rev{ImageNet-1K}, which contain 25, 100, 250, and \rev{25} samples per class, respectively. Following \citet{zhang2023gif}, we train ResNet-50 from scratch in these settings. For fine-grained classification, we follow the few-shot configurations in \citet{wang2024diffmix}, using 1, 5, and 10 samples per class from Birds and Aircraft to adapt ImageNet-pre-trained ResNet-50. For medical classification, we define data-scarce tasks with 5 and 25 samples per class on the Blood dataset, and 10 and 50 samples per class on the Skin dataset. In both cases, ResNet-50 is trained from scratch. \rev{For the more complex Semi-iNat and DomainNet datasets, we present the detailed experimental setups together with their empirical results and analyses in later sections in order to maintain a more coherent exposition.}

\textbf{Experiment Pipeline}. For each task, we follow a standardized three-stage pipeline to implement and evaluate DiffDA methods. First, under the given labeled data condition, we fine-tune the diffusion model using the specific strategy prescribed by each method, and save the resulting checkpoint. Second, guided by the real samples, each method generates synthetic data according to its own generation paradigm. Once all synthetic samples are generated and stored, the classification model is trained and evaluated following the corresponding sample utilization strategy. We report top-1 classification accuracy as the primary evaluation metric.

\textbf{Diffusion Model Fine-tuning}. We implement diffusion model finetuning using Diffusers \citep{von-platen-etal-2022-diffusers} and Accelerate \citep{accelerate}, following official documentation to apply both Textual Inversion and DreamBooth-LoRA techniques. For Textual Inversion, we set the learning rate to $5 \times 10^{-4}$ with a total batch size of 64. The model is trained for 2000 steps on natural image tasks and 500 steps on medical image tasks. DreamBooth-LoRA is then applied on top of the Textual Inversion-tuned model, using a learning rate of $5 \times 10^{-6}$, a total batch size of 64, and a LoRA rank of 8. It is trained for 1000 steps on natural images and 500 steps on medical images.

\textbf{Diffusion Sampling}. All diffusion-based sample generation is implemented using the Diffusers library. For all methods except DiffuseMix, we adopt DPMSolver++ as the noise scheduler, with the total number of diffusion steps $T = 25$. Due to the use of the InstructPix2Pix model in DiffuseMix, we follow its original configuration and apply the Euler Ancestral scheduler \citep{karras2022elucidating} with $T = 100$. All generated images have a resolution of $512 \times 512$. The augmentation ratio $M$ is set to 5 for all tasks, which means that five synthetic samples are generated per real training sample.

\textbf{Classifier Training}. We use backbone models and pretrained weights provided by the Torchvision library. All training samples, including both real and synthetic ones, are resized to a resolution of $224 \times 224$. For fine-grained datasets (Aircraft, Birds, and Semi-iNat), we apply label smoothing regularization, while for the other datasets we adopt the standard cross-entropy loss. Optimization is carried out using SGD. \rev{For each task, we select the batch size and learning rate through grid search}, and the final configurations are documented in our released codebase.

\subsection{Main Results}
\subsubsection{Conventional Classification}
We begin by presenting the results on conventional classification tasks, where the classifier is trained from scratch. Before reporting the performance metrics, we clarify several implementation details. During the sample generation stage, the image-to-image transition strength for Real Guidance, GIF, Diff-Aug, and Diff-mix is uniformly set to $s = 0.9$. For classifier training, the learning rate is set to 0.01 for \textbf{Caltech-101} and 0.1 for \textbf{CIFAR-100}, \textbf{ImageNet-100}, and \textbf{ImageNet-1K}, and the batch size is fixed to 32 in all cases. \rev{For sample utilization, we adopt the Full Concatence strategy throughout. These configurations are chosen based on our analysis of sample utilization strategies and shared hyperparameters, which is presented in detail in Sec.~\ref{subsec:4.3}}.
\begin{table}[h]
\centering
\caption{Top-1 accuracy (\%) of ResNet-50 trained from scratch on conventional classification tasks.}
\label{tab:coarse}
\begin{tabular}{ccccc}
\toprule
\multirow{2}{*}{\textbf{Methods}} & \multicolumn{4}{c}{\textbf{Datasets}} \\
\cmidrule(lr){2-5}
 & \textbf{Caltech-101} & \textbf{CIFAR-100} & \textbf{ImageNet-100} & \textbf{ImageNet-1K} \\
\midrule
Baseline & 49.95 ± 3.94 & 55.22 ± 2.90 & 65.28 ± 0.65 & 22.37 ± 0.45 \\
\midrule
Mixup    & 51.75 ± 1.79 & 58.06 ± 2.15 & 67.45 ± 1.20 & 23.54 ± 1.20 \\
CutMix   & 47.65 ± 1.39 & 56.08 ± 1.67 & 68.40 ± 1.81 & 24.08 ± 0.97 \\ 
\midrule
Real Guidance & 67.20 ± 1.57 & 66.69 ± 0.53 & 76.83 ± 0.36 & 39.07 ± 0.22 \\ 
GIF & \textbf{76.11} ± 0.33 & 68.23 ± 0.51 & \textbf{78.63} ± 0.49 & 39.52 ± 0.35 \\ 
DiffuseMix & 62.74 ± 1.47 & 58.63 ± 0.36 & 73.96 ± 0.93 & 37.35 ± 0.51 \\ 
DA-Fusion & 67.65 ± 0.40 & 66.74 ± 0.91 & 76.71 ± 0.13 & 38.87 ± 0.34 \\ 
Diff-Aug & 72.35 ± 0.96 & 68.82 ± 1.53 & 77.24 ± 0.28 & 39.45 ± 0.12 \\ 
Diff-Mix & 75.80 ± 1.16 & \textbf{71.37} ± 0.12 & 78.29 ± 0.33 & \textbf{39.93} ± 0.37 \\ 
\bottomrule
\end{tabular}
\end{table}

\begin{figure}[h]
\centering
\includegraphics[width=0.95\textwidth]{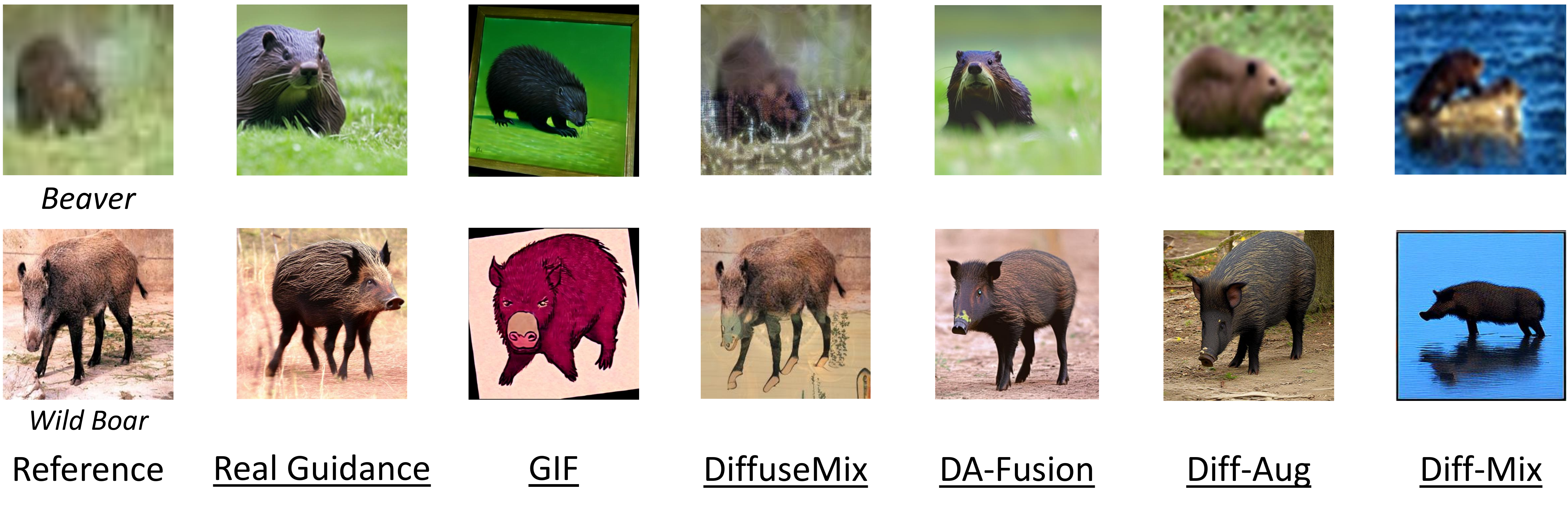}
\caption{Examples of synthetic ``Beaver'' images from CIFAR100 and ``Wild Boar'' images from ImageNet-100.}\label{fig:coarse}
\end{figure}
\vspace{-0.3cm}

\rev{Quantitative results on conventional coarse-grained classification tasks are reported in Table~\ref{tab:coarse}, and Figure~\ref{fig:coarse} further illustrates representative images generated by different DiffDA methods. For these coarse-grained datasets, diffusion models without fine-tuning already capture the underlying semantic concepts reasonably well, so the semantic fidelity of generated images is generally satisfactory across methods. As a result, the performance gap on downstream classification is mainly driven by differences in the diversity of synthetic samples. Among all approaches, GIF and Diff-Mix achieve the strongest results by explicitly enhancing diversity in different ways. GIF benefits from optimized noisy latent initialization and carefully crafted text prompts, which encourage the model to produce visually varied yet semantically consistent images. In contrast, Diff-Mix performs a generative mix-up across categories, where the foreground corresponding to the target class is combined with backgrounds from images of other classes. This leads to diverse compositions such as land animals appearing in ocean scenes, as illustrated in Figure~\ref{fig:coarse}.}

\rev{Although diffusion models can already represent these general concepts without fine-tuning, the way in which the model is fine-tuned still has a noticeable impact on the final performance. DA-Fusion only adjusts the text embeddings in order to encode target concepts, which minimally affects concepts that the diffusion model has already learned. In comparison, Diff-Aug and Diff-Mix fine-tune the denoising U-Net in addition to prompt-related components. This helps the generative model better match the statistics of the real data. The benefit is particularly evident on CIFAR-100, where there is a significant resolution gap between the real images and the untuned diffusion model, and fine-tuning effectively narrows this gap.}
\vspace{-0.3cm}
\begin{table}[h]
\centering
\caption{Top-1 accuracy (\%) of ImageNet-pre-trained ResNet-50 on few-shot fine-grained classification on the \textbf{Birds} dataset.}\label{birds}%
\begin{tabular}{ccccc}
\toprule
{\textbf{Methods}} & \textbf{1-shot} & \textbf{5-shot} & \textbf{10-shot} & {\textbf{Average}}\\
\midrule
Baseline & 16.20 ± 0.90 & 52.24 ± 0.85 & 69.94 ± 0.20 & 46.13 \\ 
\midrule
Mixup    & 13.91 ± 1.38 & 48.51 ± 0.98 & 67.40 ± 0.05 & 43.27 \\ 
CutMix  & 12.45 ± 0.91 & 35.47 ± 1.26 & 59.49 ± 0.74 & 35.80 \\ 
\midrule
Real Guidance & 17.29 ± 0.69 & 52.78 ± 0.32 & 69.94 ± 0.45 & 46.67 \\  
GIF & 17.09 ± 1.23 & 52.32 ± 0.41 & 67.76 ± 0.35 & 45.72 \\ 
DiffuseMix & 16.34 ± 1.21 & 54.48 ± 1.32 & 68.44 ± 1.51 & 46.42 \\ 
DA-Fusion & 19.21 ± 0.59 & 59.37 ± 0.81 & 70.91 ± 0.45 & 49.83 \\  
Diff-Aug & 18.37 ± 0.43 & 60.12 ± 1.02 & 71.83 ± 0.17 & 50.11 \\ 
Diff-Mix & \textbf{24.03} ± 0.53 & \textbf{63.41} ± 1.13 & \textbf{73.56} ± 1.30 & \textbf{53.67} \\ 
Diff-II & 20.34 ± 1.06 & 61.29 ± 0.76 & 72.96 ± 0.36 & 51.53 \\ 
\bottomrule
\end{tabular}
\label{tab:birds}
\end{table}
\vspace{-1.5cm}
\begin{table}[h]
\centering
\caption{Top-1 accuracy (\%) of ImageNet-pre-trained ResNet-50 on few-shot fine-grained classification on the \textbf{Aircraft} dataset.}\label{tab:aircraft}%
\begin{tabular}{ccccc}
\toprule
{\textbf{Methods}} & \textbf{1-shot} & \textbf{5-shot} & \textbf{10-shot} & {\textbf{Average}}\\
\midrule
Baseline & 10.21 ± 1.41 & 32.26 ± 1.16 & 52.64 ± 0.90 & 31.70 \\ 
\midrule
Mixup    & 7.57 ± 1.13 & 31.45 ± 0.14 & 54.12 ± 0.14 & 31.05 \\ 
CutMix  & 6.62 ± 0.74 & 24.00 ± 0.77 & 43.70 ± 0.77 & 24.77 \\ 
\midrule
Real Guidance & 11.23 ± 0.36 & 32.93 ± 0.43 & 53.72 ± 0.86 & 32.62 \\ 
GIF & 11.79 ± 0.48 & 35.12 ± 0.89 & 56.26 ± 0.43 & 34.39 \\ 
DiffuseMix & \textbf{13.03} ± 1.32 & 37.04 ± 1.38 & 57.89 ± 0.72 & 35.99 \\ 
DA-Fusion & 11.24 ± 1.75 & 33.43 ± 0.71 & 52.92 ± 0.84 & 32.53 \\ 
Diff-Aug & 11.49 ± 0.23 & 36.24 ± 0.61 & 55.34 ± 0.32 & 34.36 \\ 
Diff-Mix & 12.16 ± 1.27 & \textbf{40.59} ± 0.91 & 59.37 ± 1.02 & \textbf{37.73} \\ 
Diff-II & 11.28 ± 0.85 & 40.23 ± 0.71 & \textbf{59.62} ± 1.38 & 37.04 \\ 
\bottomrule
\end{tabular}
\end{table}
\vspace{-0.8cm}
\subsubsection{Few-Shot Fine-Grained Classification}

\rev{
The implementation details of few-shot fine-grained tasks are presented below. For SDEdit-based methods, including Real Guidance, GIF, DA-Fusion, Diff-Aug, and Diff-Mix, we use different strength values depending on whether the diffusion model is fine-tuned. For methods that rely on an untuned diffusion model (Real Guidance and GIF), we adopt a low strength of $s = 0.1$, while for methods built on a fine-tuned diffusion model (DA-Fusion, Diff-Aug, and Diff-Mix), we use a high strength of $s = 0.9$. For few-shot fine-grained tasks, all methods employ the Random Replacement strategies, where the replacement probability for each task is selected by grid search as described in Sec.~\ref{subsec:4.3}. The classifier is trained using a learning rate of 0.1 and a batch size of 256 for all tasks.
}

\begin{figure}[h]
    \centering
    \includegraphics[width=0.95\linewidth]{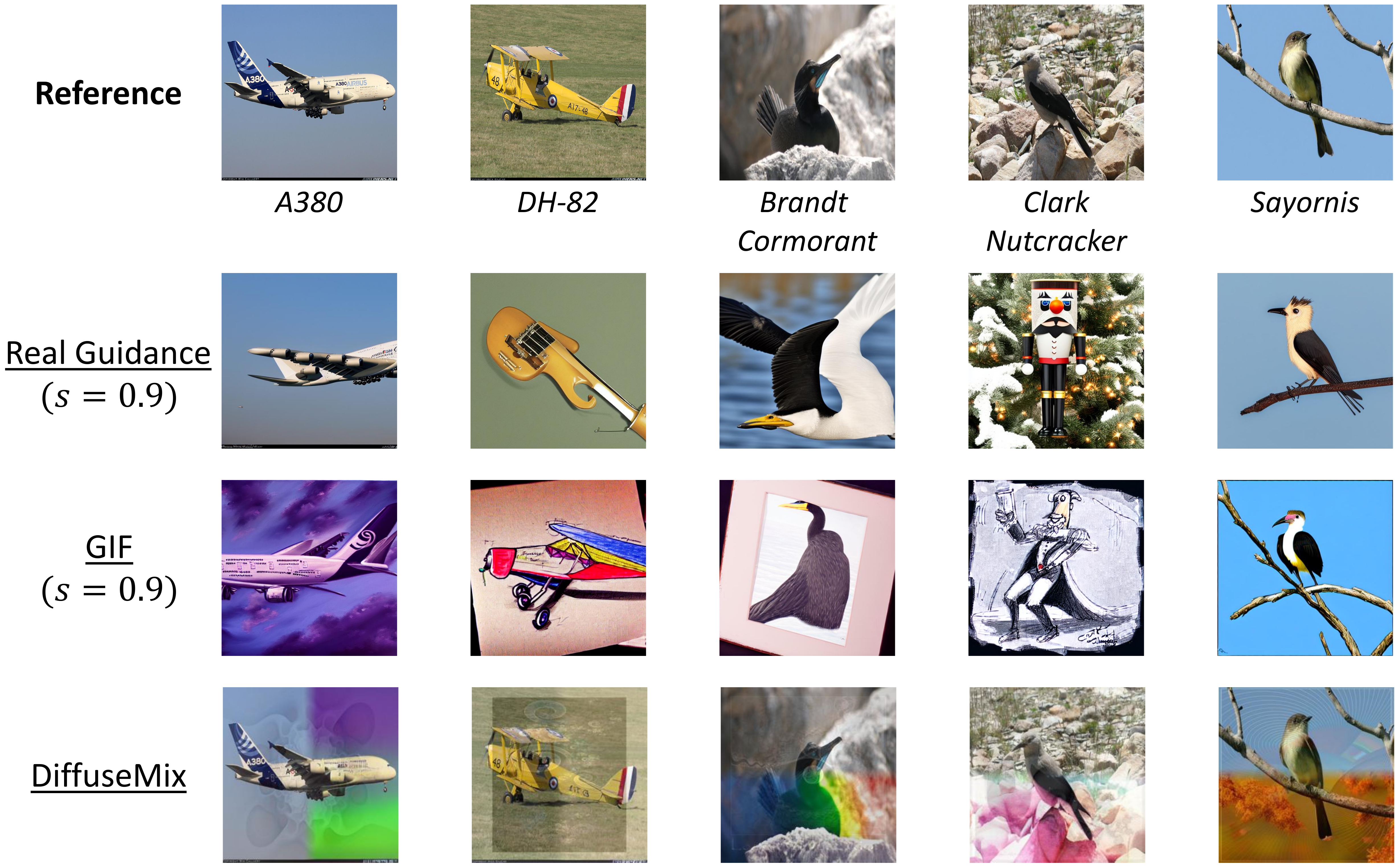}
    \caption{Examples of fine-grained concept generation with untuned diffusion models.}
    \label{fig:fine_more}
\end{figure}

\rev{Quantitative results on the \textbf{Birds} dataset are presented in Table~\ref{tab:birds}, and those on the \textbf{Aircraft} dataset are presented in Table~\ref{tab:aircraft}. To better understand these results, we first examine the qualitative examples in Fig.~\ref{fig:fine_more}, which highlight the necessity of adapting the generative model to the target fine-grained data. Due to the limited knowledge of fine-grained concepts in pre-trained diffusion models, methods that use untuned diffusion models (Real Guidance and GIF) often fail to produce semantically correct images when a high transition strength is applied. Such failures include heavily distorted or clearly irrelevant images, as well as seemingly plausible samples that nevertheless do not match the desired fine-grained semantics. This limitation is the main reason why we adopt a low strength of $s = 0.1$ for these methods when reporting their final performance. Although a low strength does not introduce substantial new information, it at least prevents the generated samples from degrading classification performance. A more detailed analysis of transition strengths is presented in Sec.~\ref{subsec:4.3}.}

\vspace{-0.3cm}
\vspace{-0.5cm}

In addition, we analyze DiffuseMix separately, since it follows a distinct image-to-image generation paradigm based on InstructPix2Pix. Rather than depending on model adaptation to preserve class labels, DiffuseMix applies style transformations that leave the underlying semantics unchanged. This design alleviates semantic distortion caused by insufficient domain-specific knowledge in the diffusion model, which can be especially helpful in challenging regimes. On the Aircraft dataset, the 1-shot setting is particularly difficult due to the high visual complexity of the categories. With only one image per class, fine-tuning the diffusion model does not provide a reliable grasp of the target concepts, so methods that rely on a fine-tuned generator (such as Diff-Mix and Diff-II) tend to underperform compared with DiffuseMix.

We further compare methods that rely on fine-tuned diffusion models. Compared with DA-Fusion, which only uses Textual Inversion, approaches that additionally fine-tune the UNet with DreamBooth-LoRA (Diff-Aug, Diff-Mix, Diff-II) consistently obtain higher accuracy, indicating a stronger capability to model new target concepts. Furthermore, although Diff-Aug, Diff-Mix, and Diff-II share the same fine-tuned diffusion model, Diff-Mix exploits inter-class image mix-up and Diff-II performs latent interpolation to generate more diverse synthetic samples. These designs lead to stronger performance than Diff-Aug and highlight the benefit of explicitly encouraging diversity during generation.

\subsubsection{Medical Classification}

We evaluate those DiffDA methods that are applicable to the medical domain using the \textbf{Blood} (blood cell images) and \textbf{Skin} (skin lesion images) datasets. Specifically, we consider five methods: Real Guidance, GIF, DA-Fusion, Diff-Aug, and Diff-Mix. Following the setting in \cite{zhang2023gif}, GIF is applied using the same fine-tuned diffusion models as Diff-Aug and Diff-Mix. Based on preliminary experiments for parameter selection, we set the $s$ to 0.3 for Real Guidance, GIF, and Diff-Aug, while $s$ is set to 0.7 for DiffMix. Since the models are trained from scratch, we adopt the same Full Concatenation strategy for all methods.

\begin{figure}[htbp]
\centering
\includegraphics[width=0.95\textwidth]{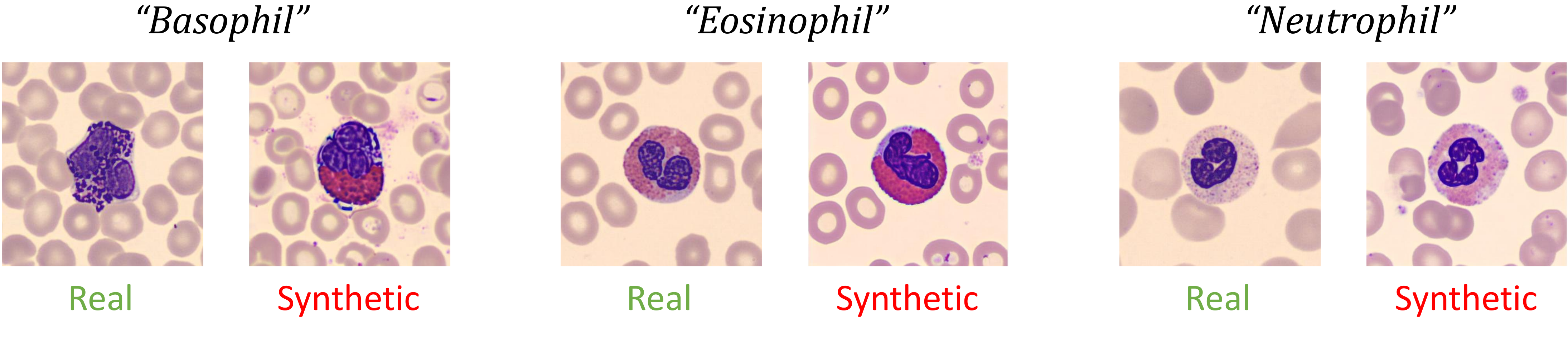}
\caption{Although diffusion models fine-tuned on a limited set of real images can generate synthetic samples with overall structural resemblance, distinguishing between different types of blood cells relies on subtle morphological cues, such as nuclear shape and cytoplasmic staining, that are too intricate for the model to capture accurately.}\label{fig:blood}
\end{figure}

Quantitative result are summarized in Table~\ref{tab:medical}. Compared to natural images, medical images often exhibit subtle inter-class differences that are crucial for semantic discrimination. These fine-grained cues are challenging for diffusion models to capture, especially under limited supervision. Therefore, adapting diffusion models to the medical domain through fine-tuning proves difficult in low-data settings, as shown in Fig.~\ref{fig:blood}. This is reflected in the results: tuning-based methods can underperform the tuning-free Real Guidance, which avoids semantic drift by using a low transition strength ($s = 0.3$) to preserve label consistency during sample generation.

\vspace{-0.6cm}

\begin{table}[h]
\centering
\caption{Top-1 accuracy (\%) of ResNet-50 trained from scratch on medical classification tasks.}
\begin{tabular}{ccccc}
\toprule
\textbf{Dataset} & \multicolumn{2}{c}{\textbf{Blood}} & \multicolumn{2}{c}{\textbf{Skin}} \\
\cmidrule(lr){1-1}  \cmidrule(lr){2-3} \cmidrule(lr){4-5}
\textbf{Number of Samples}\footnotemark[1] & \textbf{5} & \textbf{25} & \textbf{10} & \textbf{50} \\
\midrule
Baseline & 60.00 ± 1.56 & 77.67 ± 1.42 & 39.62 ± 1.75 & 50.48 ± 1.17 \\
\midrule
Mixup & 60.33 ± 7.46 & 62.17 ± 8.32 & 40.67 ± 1.41 & 47.14 ± 2.54 \\
CutMix & 60.92 ± 6.42 & 64.58 ± 10.52 & 36.86 ± 0.29 & 49.05 ± 2.31 \\
\midrule
Real Guidance & \textbf{70.67} ± 1.76 & 86.67 ± 1.53 & \textbf{42.29} ± 1.31 & 51.81 ± 0.72 \\
GIF\footnotemark[2] & 66.58 ± 3.30 & \textbf{87.25} ± 1.98 & 40.95 ± 0.16 & \textbf{51.91} ± 1.08 \\
DA-Fusion & 63.91 ± 2.32 & 80.76 ± 1.47 & 38.71 ± 2.49 & 49.00 ± 2.03 \\
Diff-Aug & 61.73 ± 3.05 & 79.06 ± 2.58 & 39.03 ± 2.84 & 51.13 ± 0.96 \\
Diff-Mix & 58.03 ± 4.67 & 70.44 ± 8.57 & 37.92 ± 3.05 & 47.85 ± 2.43 \\
\bottomrule
\end{tabular}
\vspace{-0.1cm}
\footnotetext[1]{Denoting the number of labeled samples per class.}
\vspace{-0.1cm}
\footnotetext[2]{For medical images, GIF adopts fine-tuned diffusion models.}
\label{tab:medical}
\vspace{-0.5cm}
\end{table}

\subsubsection{More Challenging Benchmarks}
\rev{First, we introduce the challenging \textbf{Semi-iNat} benchmark \citep{su2021semi_iNat} for evaluation. The dataset is constructed from the iNaturalist website and is originally proposed for evaluating semi-supervised learning algorithms. Containing 810 fine-grained species of animals, plants, and fungi, Semi-iNat naturally follows a long-tailed distribution that reflects real-world biodiversity imbalance. In this study, we use only the labeled training set, where each species contains between 5 and 80 images. }

\begin{figure}[t]
    \centering
    \includegraphics[width=0.9\linewidth]{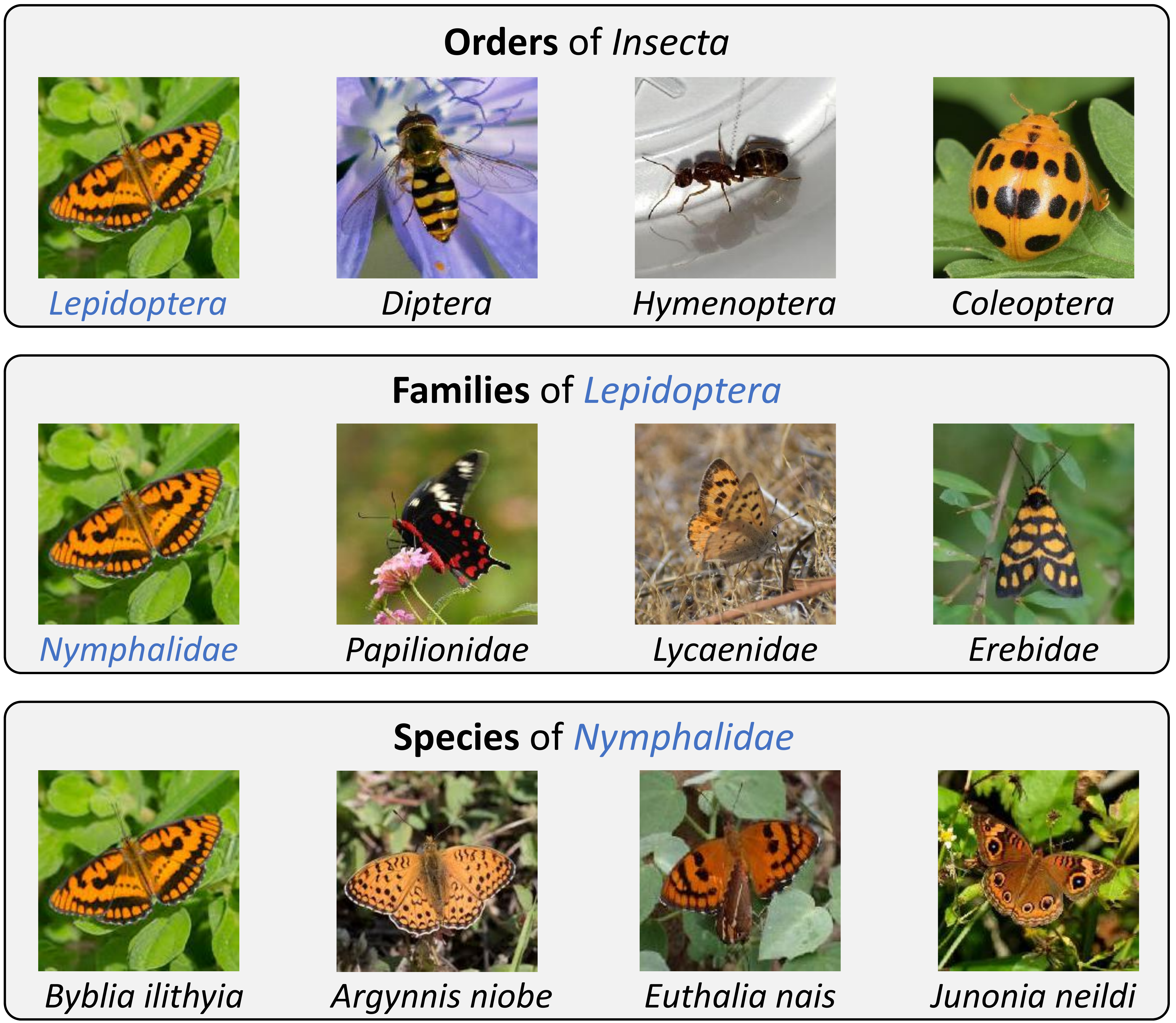}
    \caption{Illustration of hierarchical taxonomic annotations of the Semi-iNat dataset.}
    \label{fig:inat}
\end{figure}

\rev{As illustrated in Fig.~\ref{fig:inat}, each sample in Semi-iNat is annotated with seven hierarchical taxonomic levels: \textit{kingdom}, \textit{phylum}, \textit{class}, \textit{order}, \textit{family}, \textit{genus}, and \textit{species}, which allows us to analyze the impact of semantic granularity on DiffDA effectiveness. We consider three granularity levels, from fine to coarse: \textit{species}, \textit{family}, and \textit{order}. Under these settings, all 9,721 images are categorized into 810, 339, and 123 classes, respectively. We assess several DiffDA methods on this benchmark, with the results summarized in Table~\ref{tab:semantic_granularity_0}. The results indicate that overly fine semantic granularity presents substantial challenges for current DiffDA approaches. For species belonging to the same family, the subtle visual variations are often too minor for diffusion models to capture effectively, particularly when only a few training images are available per class. Under such circumstances, the generated samples frequently include semantically inaccurate details and may even deviate entirely from the intended concept, which in turn degrades downstream classification performance. }

\begin{table}[h]
\vspace{-0.5cm}
\centering
\caption{Top-1 and Top-5 accuracy (\%) of different DiffDA methods on the Semi-iNat dataset under varying semantic granularities.}
\begin{tabular}{ccccccc}
\toprule
\multirow{2}{*}{\textbf{Method}} &
\multicolumn{2}{c}{\textbf{Species}} &
\multicolumn{2}{c}{\textbf{Family}} &
\multicolumn{2}{c}{\textbf{Order}} \\
\cmidrule(lr){2-3} \cmidrule(lr){4-5} \cmidrule(lr){6-7}
 & \textbf{Top-1} & \textbf{Top-5} & \textbf{Top-1} & \textbf{Top-5} & \textbf{Top-1} & \textbf{Top-5} \\
\midrule
Baseline      & \textbf{41.88} & \textbf{64.15} & \textbf{47.85} & \textbf{70.79} & 54.96 & 79.43 \\
\midrule
Real Guidance & 38.96 & 61.89 & 44.78 & 66.63 & 53.85 & 78.27 \\
GIF           & 37.83 & 60.21 & 43.91 & 66.26 & 53.16 & 77.83 \\
Diff-Aug      & 32.91 & 55.73 & 43.96 & 66.37 & 54.58 & 79.81 \\
Diff-Mix      & 37.04 & 60.86 & 45.06 & 67.20 & \textbf{55.42} & \textbf{81.94} \\
\bottomrule
\end{tabular}
\label{tab:semantic_granularity_0}
\vspace{-0.5cm}
\end{table}

\begin{figure}[t]
    \centering
    \includegraphics[width=0.9\linewidth]{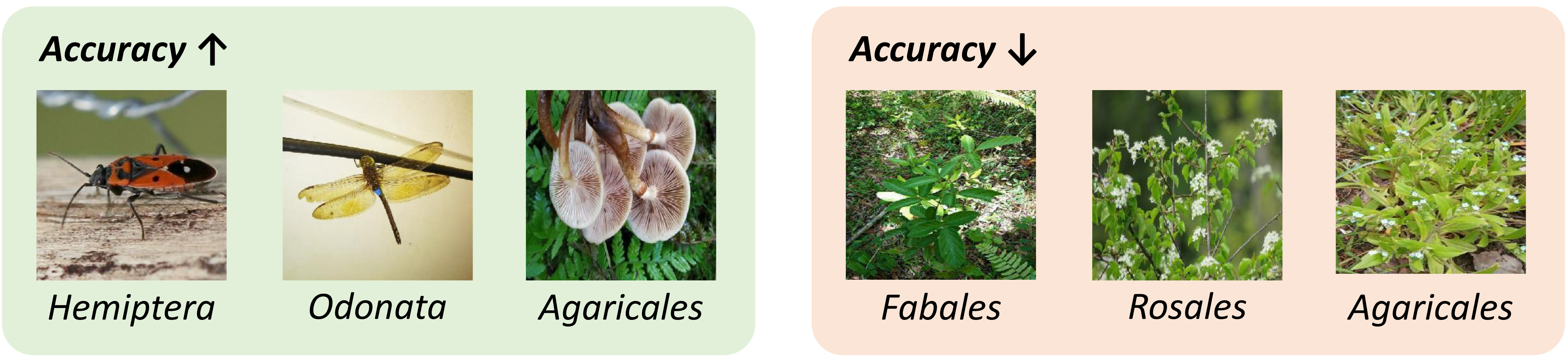}
    \caption{Examples of order-level categories from the Semi-iNat dataset that show increased or decreased classification accuracy after applying DiffDA.}
    \label{fig:inat-analysis}
\end{figure}

\rev{For classification tasks at the \textit{order} level, where inter-class differences are relatively larger, we observe that the best-performing DiffDA method finally achieves a slight overall improvement in accuracy compared with the baseline. However, a closer examination shows that not all orders benefit equally. While some categories exhibit higher accuracy after incorporating generated samples into training, others show a decline in performance. Representative examples of orders with improved and decreased accuracy are shown in Fig.~\ref{fig:inat-analysis}. From the figure, we observe that categories with improved accuracy after applying DiffDA typically contain subjects such as insects or fungi that are visually well separated from the background. In contrast, the decline in accuracy for plant categories may be attributed to the fact that the features determining their semantic labels, such as the color and shape of flowers, leaves, or stems, are less distinct in the images, making it difficult for the diffusion model to learn them effectively during fine-tuning. These findings provide a deeper understanding of the capabilities and limitations of existing DiffDA methods when applied to complex, real-world scenarios. They highlight that while DiffDA can enhance classification performance under certain conditions, its effectiveness strongly depends on the nature of the visual concepts to be learned and the clarity of the semantic cues present in the data.}

\begin{figure}[h]
\vspace{-0.5cm}
    \centering
    \includegraphics[width=0.9\linewidth]{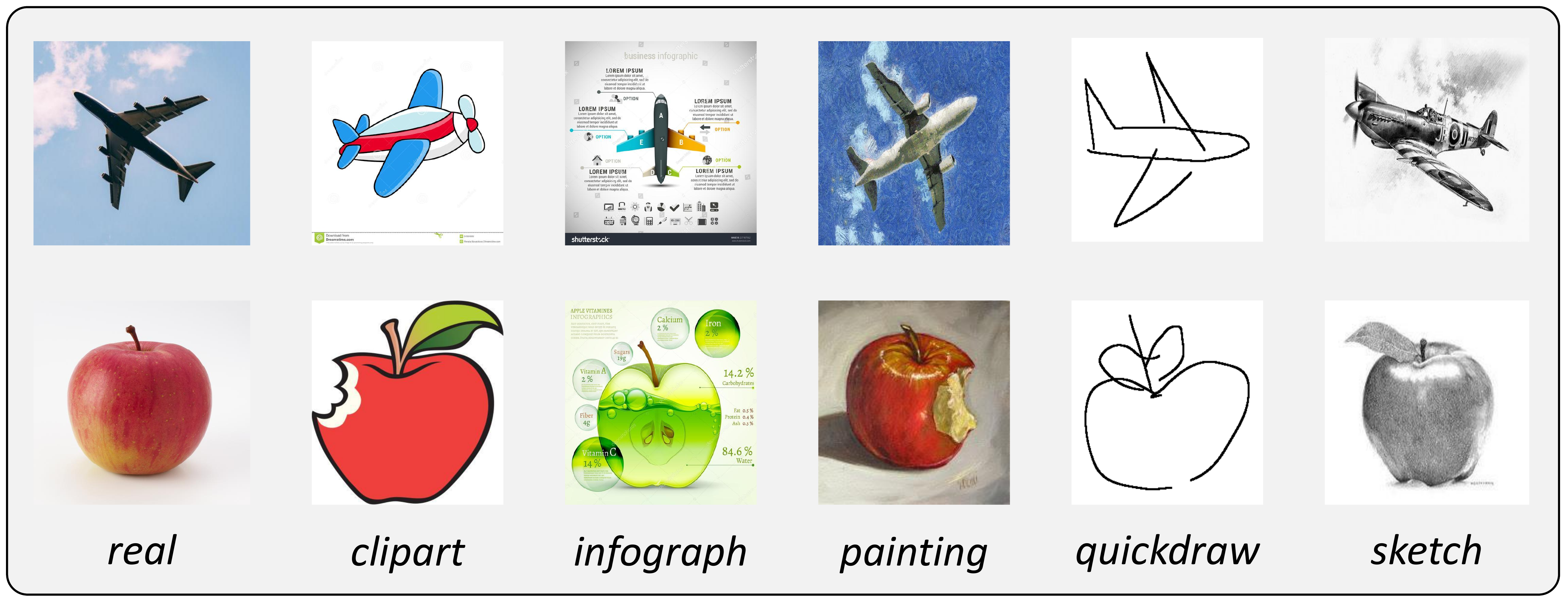}
    \caption{Visualization of multi-domain samples in the DomainNet dataset.}
    \label{fig:domainnet}
\vspace{-0.5cm}
\end{figure}

\begin{table}[h]
\centering
\caption{Top-1 accuracy (\%) of different DiffDA methods on the DomainNet dataset across six domains. Each method is trained on the \textit{real} domain and evaluated on both in-domain (\textit{real}) and out-of-domain (OOD) settings.}
\begin{tabular}{ccccccc}
\toprule
\textbf{Method} & \textbf{real} & \textbf{clipart} & \textbf{infograph} & \textbf{painting} & \textbf{quickdraw} & \textbf{sketch} \\
\midrule
Base          & 24.52 & 6.43  & 1.59 & 6.03  & 0.78 & 2.28 \\
\midrule
Real Guidance & 42.14 & 22.38 & 5.85 & 17.04 & 3.21 & 11.36 \\
GIF           & 46.81 & 26.75 & 7.30 & 22.32 & \textbf{5.09} & 18.49 \\
DiffuseMix    & 32.43 & 10.49 & 3.04 & 10.72 & 1.07 & 5.45 \\
Diff-Aug      & 48.93 & 26.93 & 8.14 & 20.72 & 3.91 & 14.81 \\
Diff-Mix      & \textbf{49.68} & \textbf{31.59} & \textbf{11.01} & \textbf{24.67} & 4.99 & \textbf{24.90} \\
\bottomrule
\end{tabular}
\label{tab:domainnet}
\vspace{-0.5cm}
\end{table}

\rev{Second, we conduct new experiments on the \textbf{DomainNet} dataset \citep{peng2019moment}, which includes six distinct domains: \textit{real}, \textit{clipart}, \textit{infograph}, \textit{painting}, \textit{quickdraw}, and \textit{sketch}, covering a total of 345 classes. Representative examples of these domains are shown in Fig.~\ref{fig:domainnet}, illustrating the large variations in visual style, texture, and abstraction level across domains. We select 20 images per class from the \textit{real} domain as the in-domain training data and evaluate multiple DiffDA methods under both in-domain (ID) testing and out-of-domain (OOD) generalization scenarios. Notably, we modify the original text prompts to the format “\texttt{a <domain-name> image of a <class-name>}” to adapt the generation process to different domains. The results, summarized in Table \ref{tab:domainnet}, show that DiffDA methods achieve clear performance gains in both ID testing and OOD generalization. However, the extent of improvement varies across domains. Considerable gains are observed on \textit{clipart}, \textit{painting}, and \textit{sketch}, likely because the diffusion model possesses better prior knowledge of these visual styles. In contrast, the generalization performance on \textit{quickdraw} remains limited, probably due to the substantial domain gap between the simple line drawings in \textit{quickdraw} and the images that the diffusion model is capable of generating. Overall, these results demonstrate that DiffDA shows strong potential to enhance the OOD generalization ability of downstream classifiers, even when the target-domain data are completely unseen during training.}

\subsection{Detailed Analysis and Discussion}
\label{subsec:4.3}
After presenting the performance of representative DiffDA methods across various benchmark tasks, we conduct a more fine-grained analysis of the main factors that drive their effectiveness. Specifically, we study: (1) different sample utilization strategies and their impact on downstream accuracy; (2) key hyperparameters involved in sample generation and utilization, including the strength of image-to-image transition ($s$) and the replacement probability ($p$) in random replacement strategies; \rev{(3) the sensitivity of DiffDA to training hyperparameters for the classifier; (4) how the size of the available training data influences the benefits of diffusion-based augmentation; (5) a qualitative analysis of generated samples that reveals typical failure modes and characteristic patterns of different methods; (6) the impact of the generative backbone by comparing different diffusion models used for data synthesis; (7) the quanlitative analysis of generative data itself;} (8) how the gains from DiffDA transfer across classification models with different capacities; and (9) the runtime of different DiffDA methods, in order to understand their practical efficiency.

\begin{center}
  \captionof{table}{Top-1 accuracy (\%) of DiffDA methods with various sample utilization strategies. For conventional classification tasks (left), classifiers are trained from scratch. For few-shot classification tasks, classifiers are fine-tuned from ImageNet-pretrained models.}
  \label{tab:util}
  \begin{minipage}[t]{0.47\textwidth}
    \centering
    \tablebodyfont
    {\setlength{\tabcolsep}{1.5pt}
    \begin{tabular}{ccccc}
    \toprule
    \multirow{2}{*}{\textbf{Methods}} & \multirow{2}{*}{\textbf{Util.}} & \multirow{2}{*}{\textbf{Caltech}} & \multirow{2}{*}{\textbf{CIFAR}} & \textbf{Birds} \\
    & & & & \textbf{(full)} \\
    \midrule
    Baseline & - & 49.95 & 55.22 & 39.71 \\
    \midrule
    \multirow{4}{*}{GIF} & FC & \textbf{76.11} & \textbf{68.23} & \textbf{50.84} \\
         &  FR & 51.43 & 31.78 & 19.98 \\
         & LRR & 56.37 & 62.79 & 42.06 \\
         & GRR & 55.62 & 62.33 & 41.78 \\
    \midrule
    \multirow{4}{*}{Diff-Aug} & FC & \textbf{71.37} & \textbf{68.82} & \textbf{66.79} \\
         &  FR & 53.46 & 56.03 & 37.45 \\
         & LRR & 55.89 & 59.89 & 53.65 \\
         & GRR & 56.04 & 59.77 & 53.73 \\
    \bottomrule
    \end{tabular}
    }
  \end{minipage}%
  \hspace{0.0 \textwidth}
  \begin{minipage}[t]{0.47\textwidth}
    \centering
    \tablebodyfont
    {\setlength{\tabcolsep}{1.5pt}
    \begin{tabular}{ccccc}
    \toprule
     \multirow{2}{*}{\textbf{Methods}} & \multirow{2}{*}{\textbf{Util.}} & \textbf{Birds} & \textbf{Birds} & \textbf{Birds} \\
     & & \textbf{(1-shot)} & \textbf{(5-shot)} & \textbf{(10-shot)} \\
    \midrule
    Baseline & - & 16.20 & 52.25 & 69.94 \\
    \midrule
    \multirow{4}{*}{DiffuseMix} & FC & 17.27 & 53.72 & 67.68 \\
         & FR & \textbf{17.64} & \textbf{55.88} & 67.14 \\
         & LRR & 16.94 & 54.48 & \textbf{68.44} \\
         & GRR & 17.07 & 54.13 & 68.27 \\
    \midrule
    \multirow{4}{*}{Diff-II} & FC & 19.06 & 60.88 & 71.07 \\
         & FR & 15.02 & 41.07 & 44.69 \\
         & LRR & \textbf{20.34} & 61.29 & \textbf{72.96} \\
         & GRR & 19.87 & \textbf{61.38} & 72.47 \\
    \bottomrule
    \end{tabular}
    }
  \end{minipage}%
  \vspace{0.5em}
\parbox{0.95\textwidth}{
\footnotesize FC = Full Concatenation, FR = Full Replacement, LRR = Local Random Replacement, GRR = Global Random Replacement \\
}
\vspace{-0.25cm}
\end{center}

\rev{\textbf{Sample Utilization Strategies.} We consider two representative scenarios: conventional classification tasks where classifiers are trained from scratch, and few-shot classification tasks where classifiers are initialized from pretrained models. For each DiffDA method investigated, we vary only the sample utilization strategy while keeping all other factors fixed. As shown in Table \ref{tab:util}, the results reveal three key findings. (1) For tasks where classifiers are trained from scratch, the Full Concatenation strategy consistently achieves the best performance on both coarse-grained datasets (such as Caltech-101 and CIFAR-100) and fine-grained datasets (such as Birds), as it allows the model to benefit from the largest amount of data. (2) For few-shot classification tasks with pretrained classifiers, seeing more images during training through the Full Concatenation strategy does not lead to further performance improvement. Instead, Random Replacement strategies achieve better performance with higher training efficiency. We also observe that the difference between Local and Global Random Replacement is marginal, suggesting that the randomization granularity has limited influence in this setting. (3) The Full Replacement strategy usually performs poorly in both settings despite providing a large training set, likely due to the complete loss of original samples. Unless measures are taken to ensure that generated samples partially preserve the original supervision signals, as in DiffuseMix, Full Replacement is not an effective strategy.}

\begin{figure}[h]
    \centering
    \begin{subfigure}[b]{0.95\linewidth}
        \centering
        \includegraphics[width=0.95\linewidth]{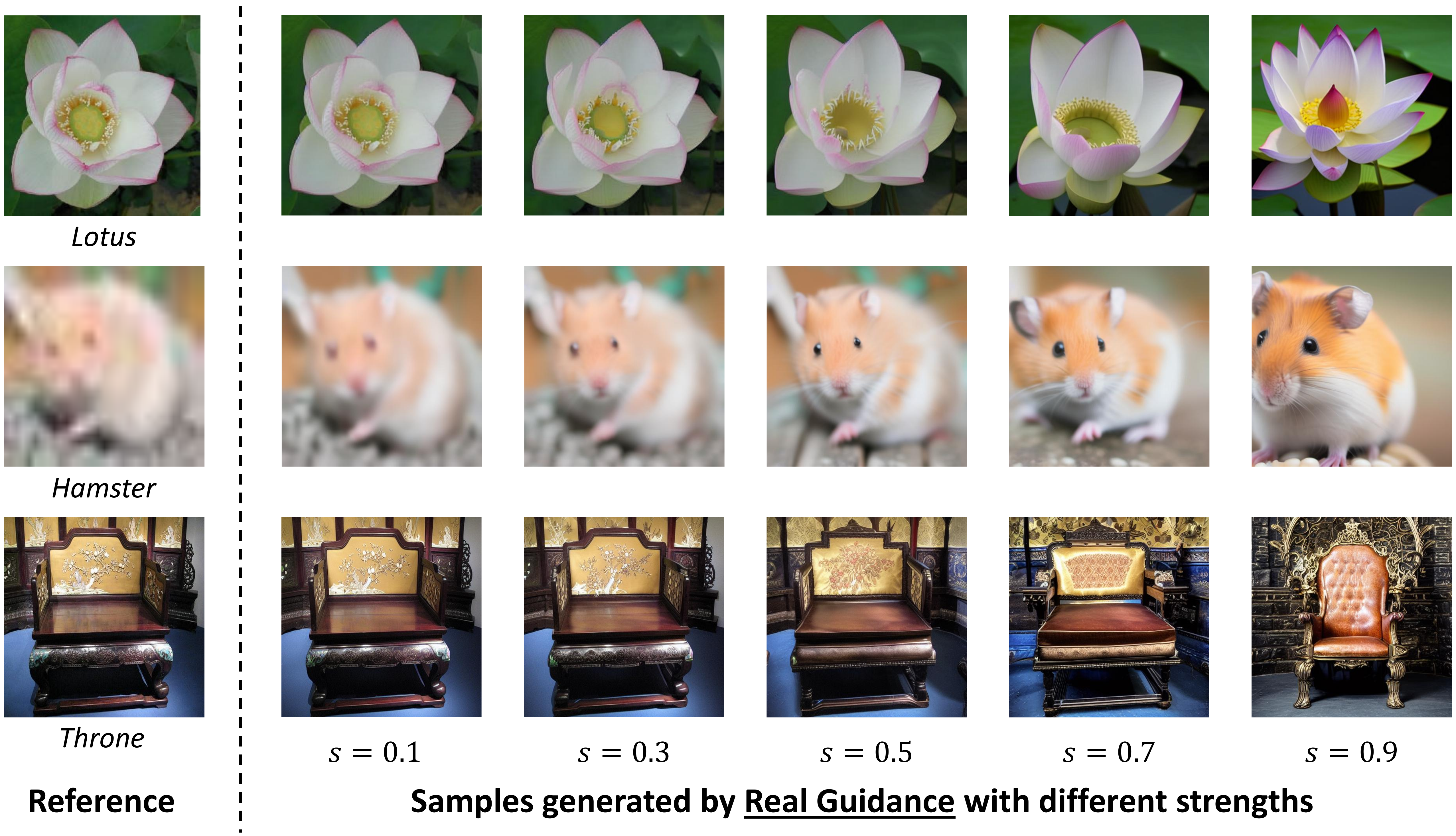}
        \caption{Examples of images with coarse-grained concepts generated by the Real Guidance method under different transition strengths. The real reference images of “Lotus”, “Hamster”, and “Throne” are taken from the Caltech-101, CIFAR-100, and ImageNet-100 datasets, respectively.} 
    \label{fig:strength_coarse}
    \end{subfigure}

    \vspace{3mm}

    \begin{subfigure}[b]{0.95\linewidth}
        \centering
        \includegraphics[width=0.95\linewidth]{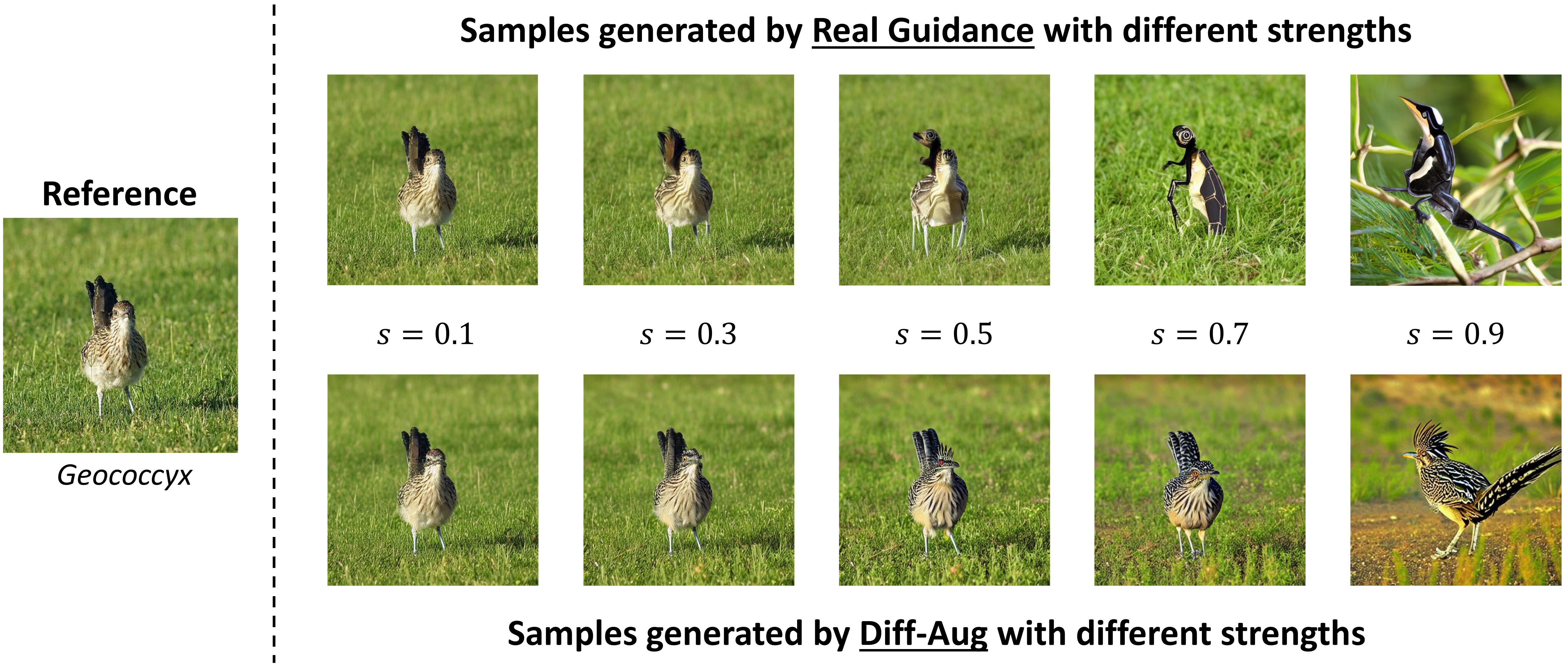}
        \caption{Examples of generated ``Geococcyx'' images from the Birds dataset. At higher transition strengths, DiffDA methods such as Real Guidance based on untuned diffusion models tend to distort fine-grained concepts.} 
        \label{fig:strength_fine}
    \end{subfigure}
\end{figure}

\begin{table}[h]
\vspace{-0.5cm}
\centering
\caption{Top-1 accuracy (\%) of Real Guidance under different image-to-image transition strengths on coarse-grained datasets.}
\label{tab:strength_coarse}
\begin{tabular}{cccccccc}
\toprule
\textbf{Datasets} & $s=0.1$ & $s=0.3$ & $s=0.5$ & $s=0.7$ & $s=0.9$ \\
\midrule
Caltech-101   & 54.96 & 56.48 & 58.70 & 64.31 & \textbf{65.38} \\
CIFAR-100     & 58.10 & 59.21 & 60.06 & 65.16 & \textbf{66.24} \\
ImageNet-100  & 66.30 & 68.16 & 70.67 & 76.70 & \textbf{77.12} \\
\bottomrule
\end{tabular}
\vspace{-0.5cm}
\end{table}

\rev{\textbf{Hyperparameters of Sample Generation and Utilization.} For mainstream DiffDA methods that follow the SDEdit paradigm, including Real Guidance, GIF, DA-Fusion, Diff-Aug, and Diff-Mix, the image-to-image transition strength is the most critical hyperparameter. It determines the extent of variation between the generated samples and their reference images, thereby controlling the diversity of the synthesized data. As illustrated in Figure \ref{fig:strength_coarse}, higher transition strengths lead to more diverse outputs compared with the original images on coarse-grained datasets, resulting in greater improvements in classification performance. These observations are further supported by the quantitative results in Table \ref{tab:strength_coarse}. Therefore, for classification tasks on coarse-grained datasets, we set the transition strength to $s = 0.9$ for all SDEdit-based DiffDA methods.}

\rev{In fine-grained classification tasks, the choice of transition strength depends on the generative backbone adopted by the DiffDA method. As shown in Figure \ref{fig:strength_fine}, for the Real Guidance method that uses an untuned diffusion model, lower strengths (e.g., $s=0.1$ and $s=0.3$) can largely preserve the semantics of the original image, although they introduce limited new information. In contrast, higher strengths tend to disrupt semantic concepts, which may lead to degraded performance. Conversely, for the Diff-Aug method, a high strength of $s=0.9$ remains the optimal choice, provided that the fine-tuned model can faithfully reproduce the target concept.}

\begin{figure}[h]
\vspace{-0.3cm}
    \centering
    \begin{subfigure}[b]{0.95\linewidth}
        \centering
        \includegraphics[width=0.95\linewidth]{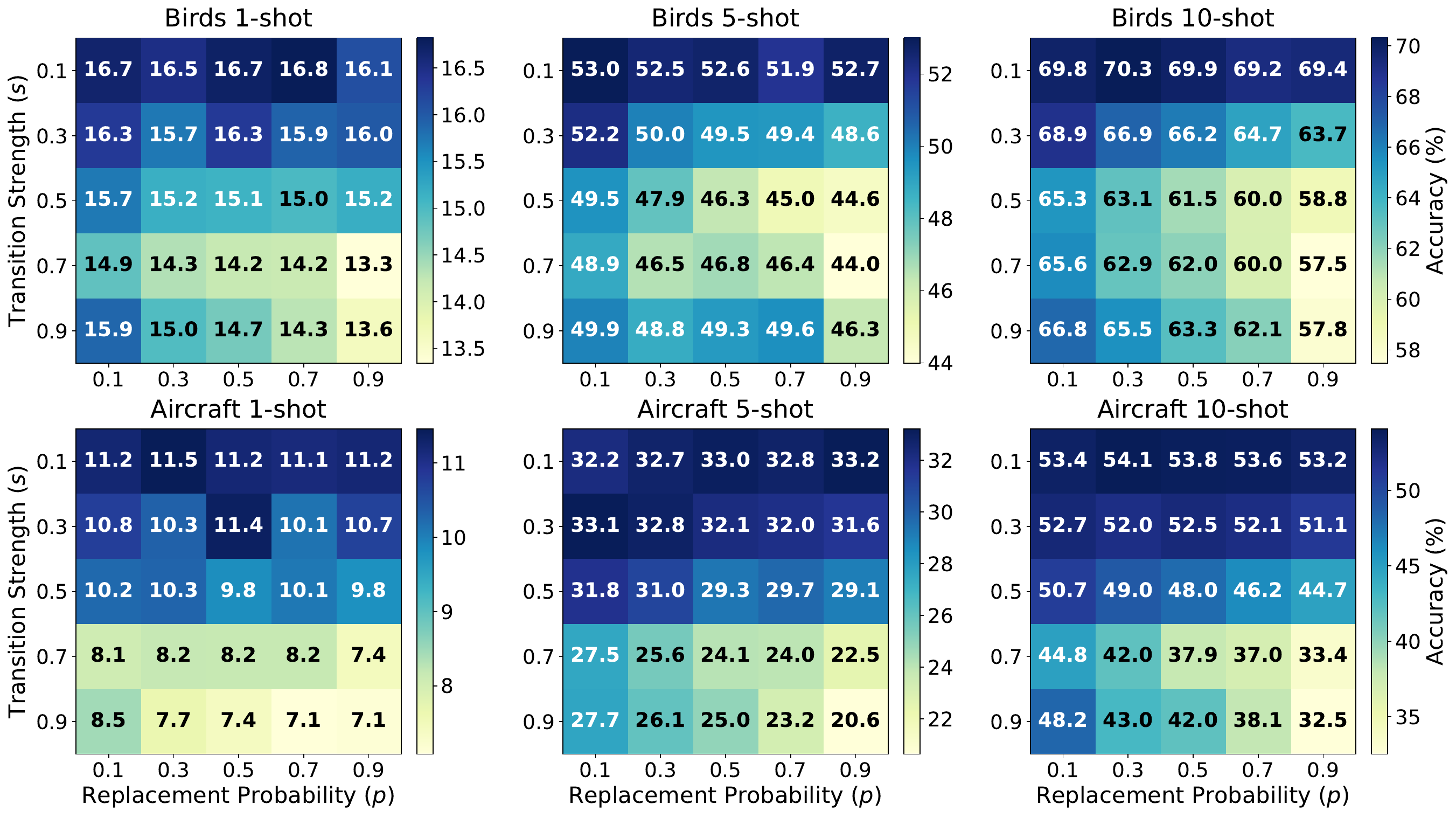}
        \caption{Grid search results of Real Guidance for different $(s, p)$ combinations.}
        \label{fig:grid_realguidance}
    \end{subfigure}


    \begin{subfigure}[b]{0.95\linewidth}
        \centering
        \includegraphics[width=0.95\linewidth]{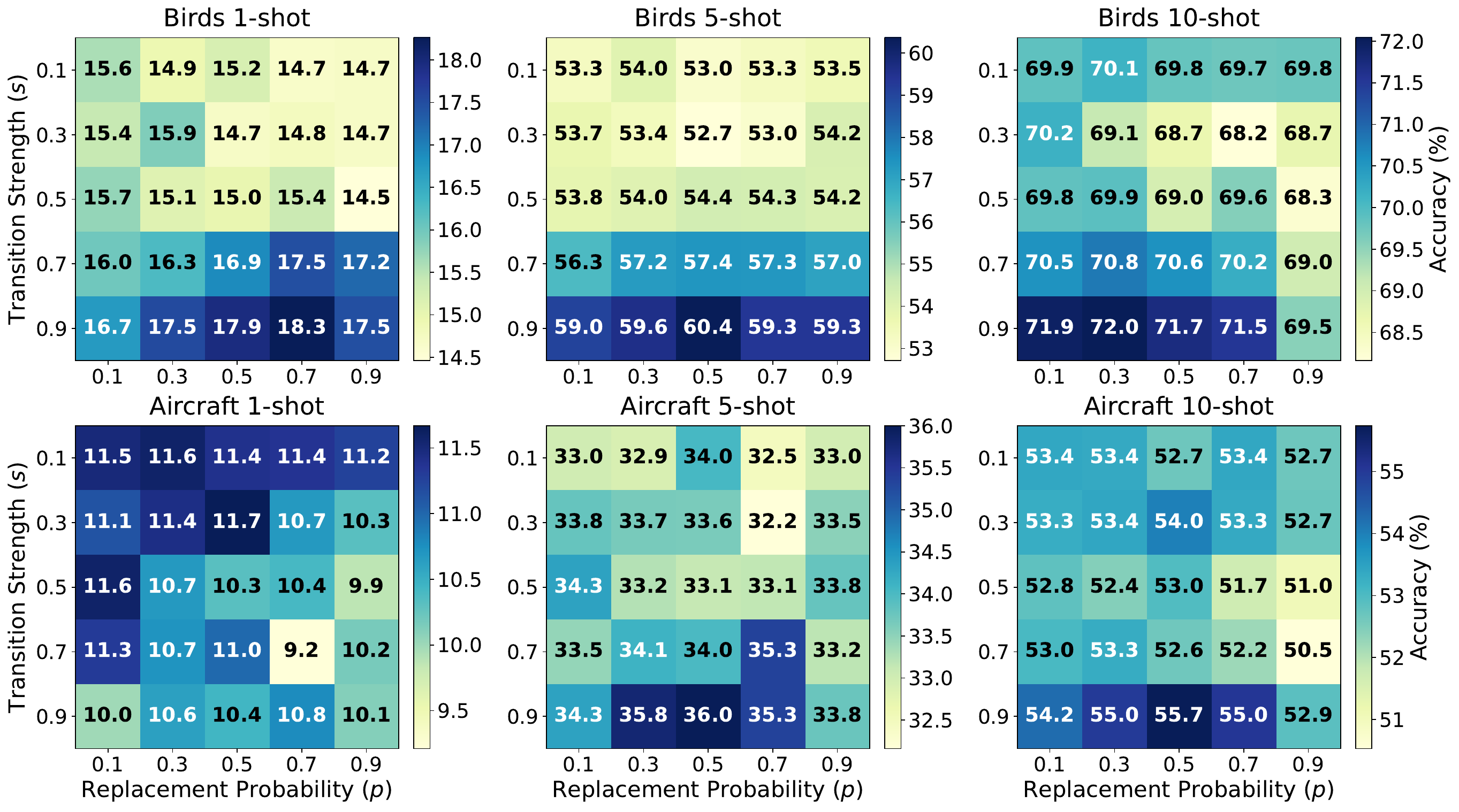}
        \caption{Grid search results of Diff-Aug for different $(s, p)$ combinations.}
        \vspace{-0.3cm}
        \label{fig:grid_diffaug}
    \end{subfigure}
    
\end{figure}

\rev{Since we adopt the Random Replacement strategies for few-shot fine-grained classification tasks, another important hyperparameter is the replacement probability $p$. To jointly analyze the impact of $s$ and $p$ on classification performance, we conduct an extensive hyperparameter grid search using Real Guidance and Diff-Aug as representative methods without and with diffusion model fine-tuning, respectively. The results of the grid search are presented in Figures \ref{fig:grid_realguidance} and \ref{fig:grid_diffaug}. For Real Guidance, setting the transition strength to $s = 0.1$ is consistently appropriate across all fine-grained classification tasks. In contrast, for Diff-Aug, a high strength of $s = 0.9$ is generally optimal, except for the Aircraft 1-shot task. We attribute this exception to the inherent complexity of the Aircraft dataset, where the diffusion model struggles to capture fine-grained concepts when only one image per class is available. The grid search results further demonstrate that the performance is substantially less sensitive to $p$ than to $s$. While the optimal value of $p$ varies across tasks, selecting a default value around $p = 0.5$ can serve as a reasonable choice.}

\begin{figure}[h]
    \centering
    \includegraphics[width=0.95\linewidth]{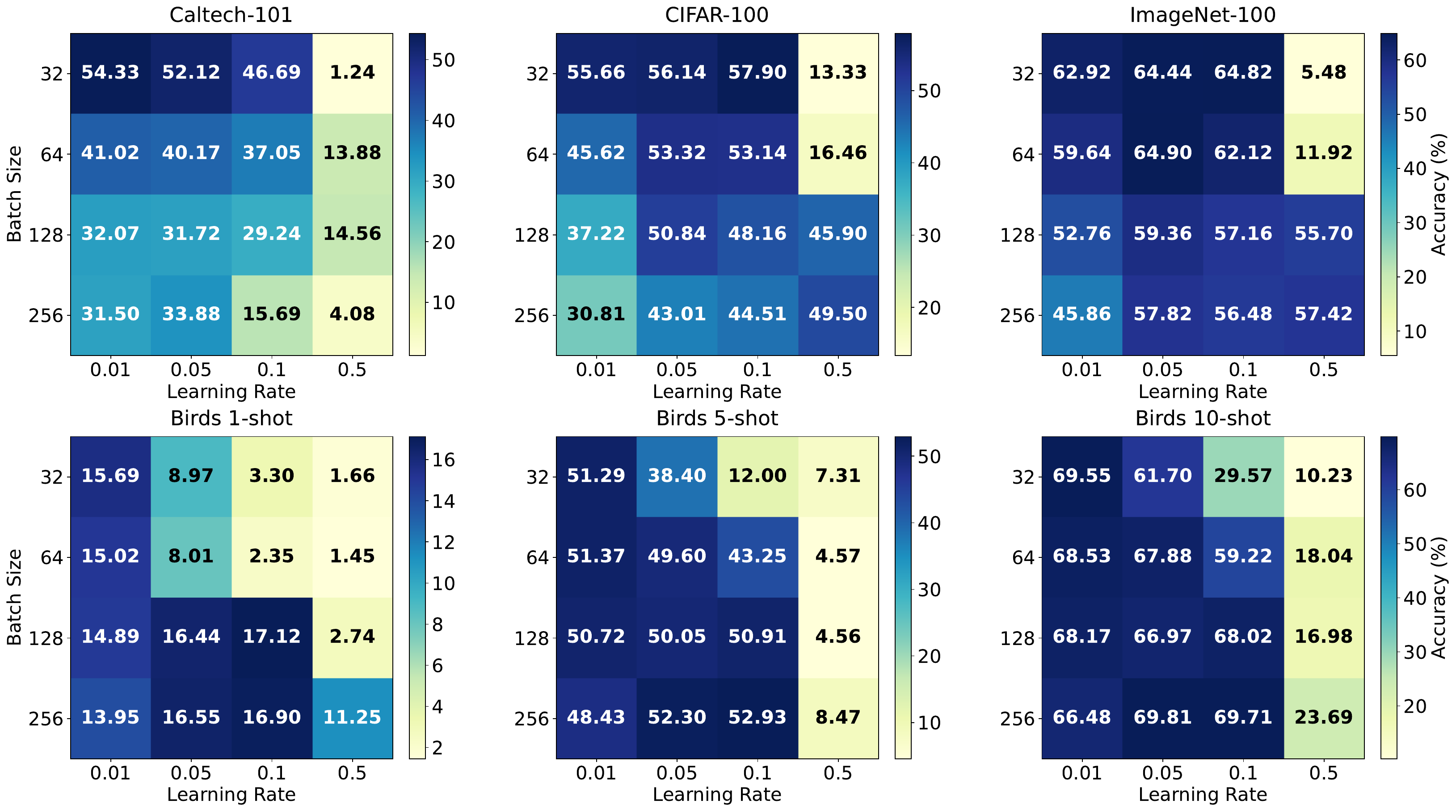}
    \caption{Grid search results for classifier hyperparameters, including batch size and learning rate.}
    \label{fig:grid_classifier}
    \vspace{-0.5cm}
\end{figure}

\rev{\textbf{Hyperparameters for Training Classifiers.} In addition to the hyperparameters related to the generation process, we also analyze two key hyperparameters for classifier training, namely the batch size and learning rate. Specifically, the batch size varies among $\{32, 64, 128, 256\}$, and the learning rate varies among $\{0.01, 0.05, 0.1, 0.5\}$. To ensure the generality of the parameter selection, we conduct this analysis by training and testing classifiers on the original datasets, which correspond to the “Baseline” results in the main experiments. The results are presented in Figure \ref{fig:grid_classifier}. Guided by these results, in the final evaluation of all DiffDA methods, we set the batch size to 32 for conventional classification tasks on coarse-grained datasets, with the learning rate fixed at 0.01 for Caltech-101 and 0.1 for CIFAR-100, ImageNet-100, and ImageNet-1K. For few-shot fine-grained classification tasks on the Birds and Aircraft datasets, we use a batch size of 256 and a learning rate of 0.1 uniformly.}

\rev{\textbf{Training Data Size.} The impact of training data size on the effectiveness of DiffDA has been partially demonstrated by the results of the few-shot fine-grained classification tasks (Table \ref{tab:birds} and Table \ref{tab:aircraft}). Given the same label space and test data, the number of shots, that is, the number of training samples per class, determines the difficulty of the fine-grained classification task. On the Birds dataset, the 1-shot and 5-shot tasks are more challenging, and DiffDA leads to substantial performance improvements in these settings. In contrast, for the 10-shot task, where the task is relatively easier, the performance gain becomes less pronounced. The training data size also affects the difficulty of fine-tuning diffusion models. In the Aircraft dataset, the 1-shot task is particularly challenging due to the intrinsic complexity of the data. With only one image available per class, the diffusion model cannot effectively learn fine-grained concepts through fine-tuning, which causes methods relying on fine-tuned models (such as Diff-Mix and Diff-II) to underperform compared with DiffuseMix, which does not require model fine-tuning but preserves structural consistency through specific generation designs. However, in the 5-shot and 10-shot tasks, methods based on fine-tuned models regain their advantage. These observations suggest that the choice of DiffDA method should consider not only the training data size but also the characteristics of the dataset itself.}

\rev{Furthermore, we investigated the effect of training data size on the impact of DiffDA in conventional classification settings, where classifiers are trained from scratch. Specifically, we partitioned the Caltech-101 dataset into subsets containing $\{20\%, 40\%, 60\%, 80\%, 100\%\}$ of the original training data and generated additional synthetic samples corresponding to $1\times$, $5\times$, $10\times$, and $20\times$ the size of each subset. This setup allows for a comprehensive evaluation of how both the amount of real data and the amount of generated data influence classification performance. We evaluate the GIF method on these tasks, with the results presented in Table \ref{tab:data_scale}. Two main observations can be drawn from the results. (1) For a fixed subset of real data, adding more generated samples consistently improves classification performance, but the efficiency of such improvement diminishes as the number of synthetic samples increases. Expanding the total data size from $(1+5)\times$ to $(1+20)\times$ yields much smaller gains compared with expanding from $1\times$ to $(1+5)\times$. (2) Real data remain substantially more valuable than synthetic data. For example, using 100\% of the real data without any augmentation achieves an accuracy of 54.33\%, whereas expanding only 20\% of the the real data to $(1+20)\times$, a total dataset four times larger than the full real dataset, achieves only 49.58\%.}

\begin{table}[h]
\vspace{-0.5cm}
\centering
\caption{Top-1 accuracy (\%) of the GIF method on conventional classification tasks of the Caltech-101 dataset with various combinations of real and generated data sizes.}
\begin{tabular}{cccccc}
\toprule
\multirow{2}{*}{\textbf{Real Data Size}} & \multicolumn{5}{c}{\textbf{Expansion Ratio}} \\
\cmidrule(lr){2-6}
 & \textbf{1$\times$} & \textbf{(1+1)$\times$} & \textbf{(1+5)$\times$} & \textbf{(1+10)$\times$} & \textbf{(1+20)$\times$} \\
\midrule
20\%  & 14.27 & 21.64 & 34.96 & 41.98 & 49.58 \\
40\%  & 24.19 & 31.27 & 54.56 & 61.42 & 66.06 \\
60\%  & 34.05 & 47.93 & 67.08 & 69.58 & 74.05 \\
80\%  & 40.68 & 55.30 & 72.08 & 74.90 & 78.53 \\
100\% & 54.33 & 60.23 & 76.49 & 79.21 & 82.04 \\
\bottomrule
\end{tabular}
\label{tab:data_scale}
\vspace{-0.5cm}
\end{table}

\begin{figure}[h]
    \centering
    \includegraphics[width=0.95\linewidth]{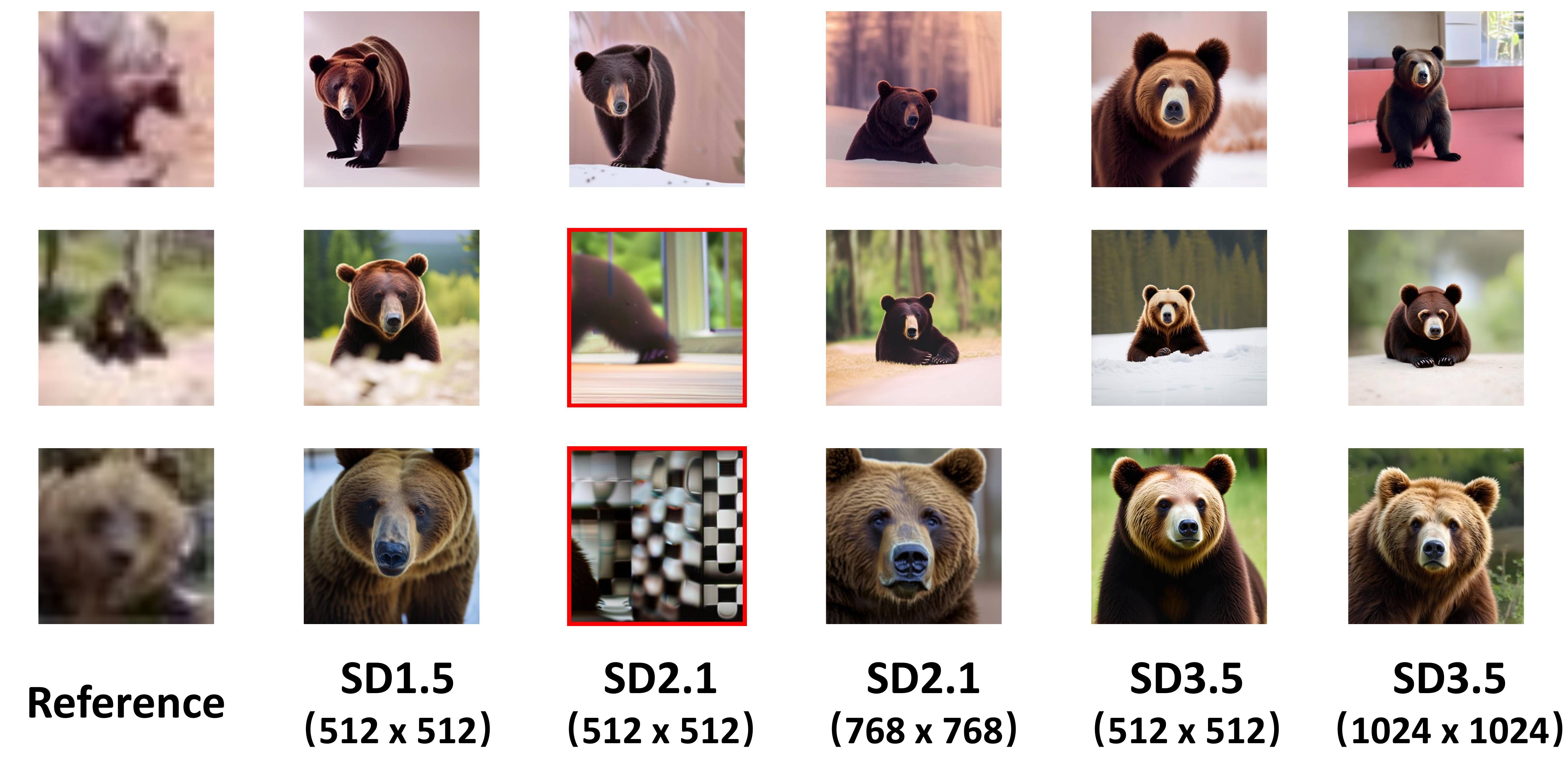}
    \caption{Examples of ``bear'' images from the CIFAR-100 dataset generated by the Real Guidance method with different generative backbones at various resolutions. Failure cases of SD2.1 caused by resolution mismatch are indicated with red boxes.}
    \label{fig:new_sd_coarse}
    \vspace{-0.5cm}
\end{figure}

\begin{figure}[h]
    \centering
    \includegraphics[width=0.75\linewidth]{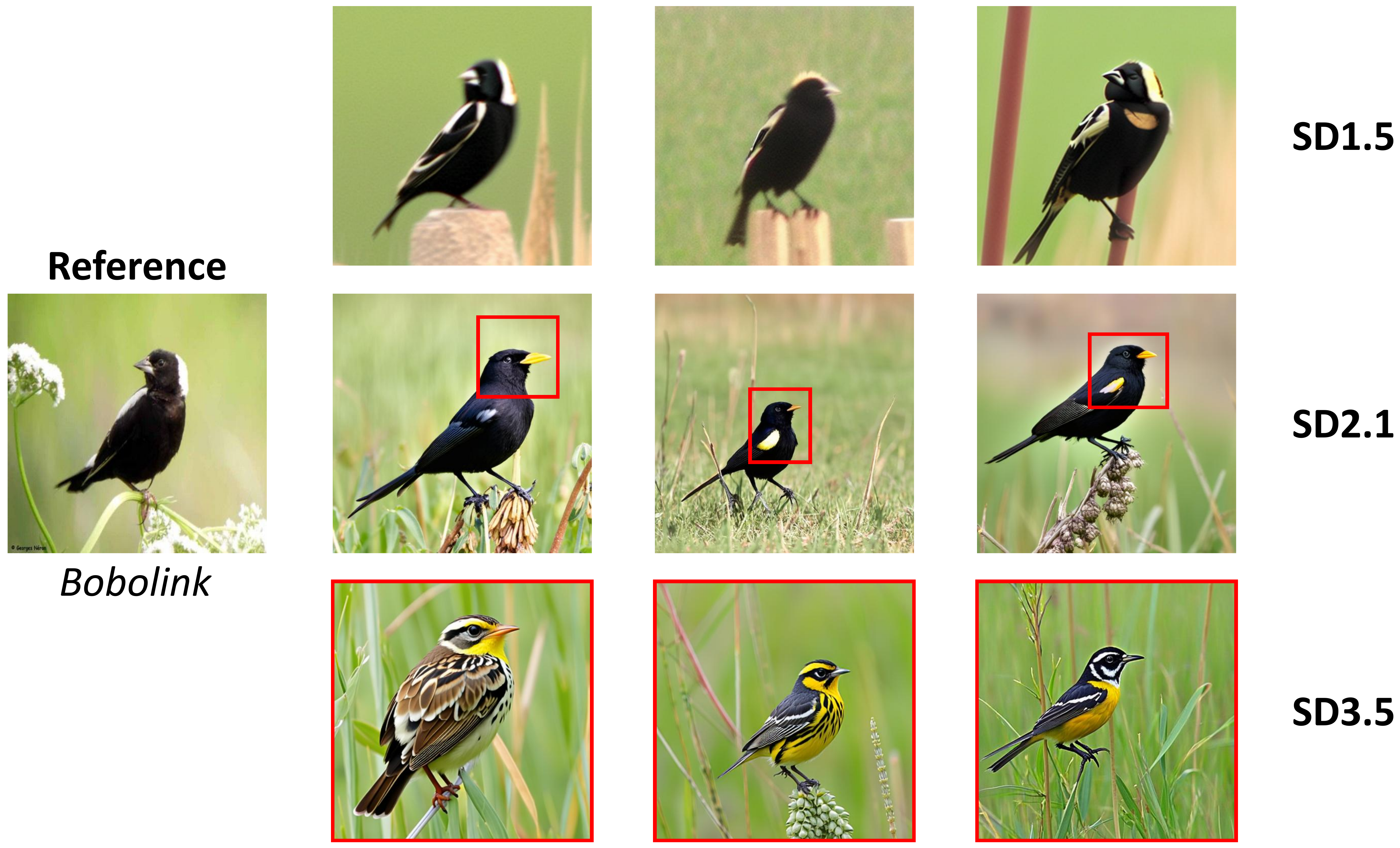}
    \caption{Examples of ``bobolink'' images from the Birds dataset generated by the Diff-Aug method with different generative backbones. Incorrectly synthesized details and complete failure cases are indicated with red boxes.}
    \label{fig:new_sd_fine}
    \vspace{-0.5cm}
\end{figure}

\rev{\textbf{Generative Backbones.} We conduct an extensive analysis to examine whether more advanced generative foundation models, such as Stable Diffusion 2.1 and 3.5 versions, could further improve the effectiveness of diffusion-based data augmentation for image recognition. To clearly isolate the influence of generative backbones on performance, we first adopt the simplest Real Guidance method with different pre-trained backbones and evaluate it on coarse-grained datasets, Caltech-101 and CIFAR-100. The results are presented in Table \ref{tab:sd_coarse}. When the output resolution of the three backbones is unified to $512 \times 512$, the performance of newer backbones decreases compared with SD1.5, with SD2.1 showing a particularly notable drop. We attribute this phenomenon to the mismatch between the resolution of the generated images and that of the training data used for the corresponding generative models, since SD2.1 is mainly trained on $768 \times 768$ images, whereas SD3.5 is trained with images of $1024 \times 1024$ or higher resolution. In addition, the UNet architecture employed in SD2.1 is less flexible in supporting varying resolutions compared with the DiT architecture used in SD3.5. When we generate images at $768 \times 768$ and $1024 \times 1024$ resolutions using SD2.1 and SD3.5 respectively, the final classification performance improves, but the advantage over SD1.5 remains marginal. We present qualitative examples in Fig. \ref{fig:new_sd_coarse}, which illustrate that more advanced generative backbones can produce samples with higher visual quality, such as finer details in the bear’s face. However, this improvement in visual quality contributes little to the final classification performance. Moreover, generating higher-resolution samples substantially increases the computational cost of the generation process.}

\rev{Furthermore, we fine-tuned the SD2.1 and SD3.5 models following the same procedure as for SD1.5, and evaluated the performance of the Diff-Aug and Diff-Mix methods using these fine-tuned models. The results are presented in Table \ref{tab:sd_tuned}. We observe that, for coarse-grained classification tasks, different generative backbones yield comparable performance. However, for fine-grained tasks, more advanced generative backbones may lead to a noticeable degradation in performance. We provide qualitative examples in Fig. \ref{fig:new_sd_fine} to illustrate this phenomenon. For SD1.5, although the fine-tuned model may generate images that appear somewhat blurry, it is still able to capture fine-grained visual cues essential for bird species recognition, such as the color of head and back feathers, as well as the shape and color of the beak. In contrast, more advanced backbones, despite producing visually higher-quality images, tend to lose these subtle yet critical details. As a result, incorporating such generated samples into classifier training can even degrade downstream performance.}

\rev{In summary, more advanced generative backbones do not necessarily lead to performance improvements in the DiffDA setting, at least for the existing methods evaluated in this study. It will be interesting to explore how to better leverage the stronger generative priors provided by these advanced models to enhance data augmentation effectiveness.}

\begin{table}[h]
\vspace{-0.5cm}
\centering
\caption{Top-1 accuracy (\%) of Real Guidance with different generative backbones at various resolutions.}
\begin{tabular}{cccccc}
\toprule
\multirow{2}{*}{\textbf{Dataset}} & \multicolumn{1}{c}{\textbf{SD1.5}} & \multicolumn{2}{c}{\textbf{SD2.1}} & \multicolumn{2}{c}{\textbf{SD3.5}} \\
\cmidrule(lr){2-2} \cmidrule(lr){3-4} \cmidrule(lr){5-6}
 & \(512 \times 512\) & \(512 \times 512\) & \(768 \times 768\) & \(512 \times 512\) & \(1024 \times 1024\) \\
\midrule
Caltech-101 & 65.38 & 63.34 & 65.27 & 64.76 & \textbf{65.44} \\
CIFAR-100   & 66.24 & 59.87 & \textbf{66.86} & 65.10 & 66.17 \\
\bottomrule
\end{tabular}
\label{tab:sd_coarse}
\vspace{-0.5cm}
\end{table}

\begin{table}[h]
\vspace{-1.2cm}
\centering
\caption{Top-1 accuracy (\%) of Diff-Aug and Diff-Mix with different generative backbones.}
\begin{tabular}{ccccc}
\toprule
\textbf{Method} & \textbf{Caltech-101} & \textbf{CIFAR-100} & \textbf{Birds (5-shot)} & \textbf{Birds (10-shot)} \\
\midrule
Diff-Aug (SD1.5) & \textbf{71.67} & 67.73 & \textbf{60.36} & \textbf{71.71} \\
Diff-Aug (SD2.1) & 71.27 & \textbf{68.19} & 48.73 & 64.01\\
Diff-Aug (SD3.5) & 71.52 & 66.97 & 44.65 & 58.13 \\
\midrule
Diff-Mix (SD1.5) & \textbf{75.62} & 71.45 & \textbf{62.96} & \textbf{73.26} \\
Diff-Mix (SD2.1) & 75.35 & \textbf{71.57} & 51.14 & 67.75 \\
Diff-Mix (SD3.5) & 75.21 & 70.64 & 46.10 & 60.27 \\
\bottomrule
\end{tabular}
\label{tab:sd_tuned}
\vspace{-0.85cm}
\end{table}

\rev{\textbf{Quantitative Analysis of Generated Data Itself.} Since our work focuses on understanding to what extent diffusion-generated samples can improve classifier performance in data-scarce scenarios, downstream classification accuracy indeed serves as the gold standard for evaluating different DiffDA methods. Nevertheless, we have also conducted a quantitative analysis of the generated data itself. We evaluated the images produced by various DiffDA methods using two commonly used metrics in generative model research: (1) Fréchet Inception Distance (FID) \citep{FID}, which measures the difference between the distribution of generated samples and real samples. (2) Precision and Recall \citep{ImprovedPrecisionRecall}, which respectively measure how closely the generated data match the true data distribution and how well the generated data cover the true data distribution.} 

\begin{table}[h]
\vspace{-0.5cm}
\centering
\caption{Quantitative analysis of images generated by different DiffDA methods on coarse-grained datasets.}
\label{tab:metric-coarse}
\setlength{\tabcolsep}{3.5pt}
\begin{tabular}{ccccccccc}
\toprule
\multirow{2}{*}{\textbf{Methods}} &
\multicolumn{4}{c}{\textbf{Caltech-101}} &
\multicolumn{4}{c}{\textbf{CIFAR-100}} \\
\cmidrule(lr){2-5} \cmidrule(lr){6-9} 
 & \textbf{FID$\downarrow$} & \textbf{Prec.$\uparrow$} & \textbf{Rec.$\uparrow$} & \textbf{\underline{Acc.}$\uparrow$}
 & \textbf{FID$\downarrow$} & \textbf{Prec.$\uparrow$} & \textbf{Rec.$\uparrow$} & \textbf{\underline{Acc.}$\uparrow$} \\
\midrule
Baseline & - & - & - & 0.50 & - & - & - & 0.55  \\
\midrule
Real Guidance & 42.79 & 0.38 & 0.28 & 0.67 & 84.45 & 0.03 & 0.17 & 0.67  \\
GIF & 54.81 & 0.40 & 0.33 & \textbf{0.76} & 113.22 & 0.01 & 0.10 & 0.68  \\
DiffuseMix & 41.22 & 0.82 & 0.87 & 0.63 & 149.97 & 0.01 & 0.09 & 0.59  \\
DA-Fusion & 41.99 & 0.38 & 0.29 & 0.68 & 83.54 & 0.03 & 0.18 & 0.67  \\
Diff-Aug & 25.71 & 0.75 & 0.32 & 0.72 & 23.90 & 0.46 & 0.51 & 0.69  \\
Diff-Mix & 27.45 & 0.60 & 0.43 & 0.76 & 20.03 & 0.37 & 0.58 & \textbf{0.71}  \\
\bottomrule
\end{tabular}%
\vspace{-0.9cm}
\end{table}

\begin{table}[h]
\vspace{-0.5cm}
\centering
\setlength{\tabcolsep}{4.5pt} 
\caption{Quantitative analysis of images generated by different DiffDA methods on the Birds dataset.}
\label{tab:metric-birds}
\begin{tabular}{ccccccccc}
\toprule
\multirow{2}{*}{\textbf{Methods}} & \multicolumn{4}{c}{\textbf{5-shot}} & \multicolumn{4}{c}{\textbf{10-shot}} \\
\cmidrule(lr){2-5} \cmidrule(lr){6-9} 
 & \textbf{FID$\downarrow$} & \textbf{Prec.$\uparrow$} & \textbf{Rec.$\uparrow$} & \textbf{\underline{Acc.}$\uparrow$} & \textbf{FID$\downarrow$} & \textbf{Prec.$\uparrow$} & \textbf{Rec.$\uparrow$} & \textbf{\underline{Acc.}$\uparrow$} \\
 \midrule
Baseline & - & - & - & 0.52 & - & - & - & 0.70 \\
 \midrule
Real Guidance & 6.54 & 0.99 & 0.05 & 0.53 & 4.32 & 0.99 & 0.10 & 0.70\\
GIF            & 16.20 & 0.73 & 0.19 & 0.52 & 15.58 & 0.71 & 0.22 & 0.68 \\
DiffuseMix     & 28.69 & 0.63 & 0.27 & 0.54 & 28.79 & 0.62 & 0.27 & 0.68\\
DA-Fusion      & 17.83 & 0.41 & 0.23 & 0.59 & 15.80 & 0.32 & 0.24 & 0.71\\
Diff-Aug       & 13.27 & 0.46 & 0.29 & 0.60 & 13.05 & 0.44 & 0.26 & 0.72\\
Diff-Mix       & 13.04 & 0.43 & 0.36 & \textbf{0.63} & 12.90 & 0.41 & 0.33 & \textbf{0.74}\\
Diff-II        & 14.16 & 0.45 & 0.26 & 0.61 & 13.75 & 0.46 & 0.28 & 0.73\\
\bottomrule
\end{tabular}
\vspace{-0.9cm}
\end{table}

\begin{table}[h]
\vspace{-0.5cm}
\centering
\setlength{\tabcolsep}{4.5pt} 
\caption{Quantitative analysis of images generated by different DiffDA methods on the Aircraft dataset.}
\label{tab:metric-aircraft}
\begin{tabular}{ccccccccc}
\toprule
\multirow{2}{*}{\textbf{Methods}} & \multicolumn{4}{c}{\textbf{5-shot}} & \multicolumn{4}{c}{\textbf{10-shot}} \\
\cmidrule(lr){2-5} \cmidrule(lr){6-9} 
 & \textbf{FID$\downarrow$} & \textbf{Prec.$\uparrow$} & \textbf{Rec.$\uparrow$} & \textbf{\underline{Acc.}$\uparrow$} & \textbf{FID$\downarrow$} & \textbf{Prec.$\uparrow$} & \textbf{Rec.$\uparrow$} & \textbf{\underline{Acc.}$\uparrow$} \\
 \midrule
 Baseline & - & - & - & 0.32 & - & - & - & 0.53 \\
 \midrule
Real Guidance & 5.95 & 0.99 & 0.04 & 0.33 &  3.42 & 0.99 & 0.07 & 0.54\\
GIF            & 23.14 & 0.39 & 0.20 & 0.35 & 22.15 & 0.38 & 0.22 & 0.56\\
DiffuseMix     & 49.52 & 0.31 & 0.19 & 0.37 & 49.90 & 0.33 & 0.37 & 0.58\\
DA-Fusion      & 14.30 & 0.33 & 0.50 & 0.33 & 15.62 & 0.37 & 0.27 & 0.53\\
Diff-Aug       & 7.91 & 0.37 & 0.44 & 0.36 & 7.15 & 0.38 & 0.42 & 0.55\\
Diff-Mix       & 7.93 & 0.35 & 0.48 & \textbf{0.41} & 7.01 & 0.37 & 0.45 & 0.59\\
Diff-II        & 12.02 & 0.35 & 0.40 & 0.40 & 11.84 & 0.34 & 0.44 & \textbf{0.60}\\
\bottomrule
\end{tabular}
\vspace{-0.5cm}
\end{table}

\rev{In Tables~\ref{tab:metric-coarse}, \ref{tab:metric-birds}, and \ref{tab:metric-aircraft}, we report the FID and Precision and Recall scores of the images generated by different DiffDA methods across various tasks. For reference, we also include the corresponding downstream classification accuracies. From the results, we observe that conventional metrics for evaluating generative models do not accurately reflect the effectiveness of DiffDA methods. For example, on the CIFAR-100 task, Real Guidance and GIF, which use untuned diffusion models, show relatively poor quantitative scores. This happens because the original CIFAR-100 images have a resolution of $32 \times 32$, while the generated samples are at $512 \times 512$, which leads to a large mismatch between the real and generated data distributions. Nevertheless, the high-resolution generated images actually help the classifier learn better in the downstream task. Another example appears in the fine-grained datasets, where Real Guidance achieves very low FID scores, indicating high visual similarity between generated and real samples. However, this is mainly due to the use of a low transition strength ($s=0.1$), which makes the generated samples almost identical to the real ones without introducing new and useful variations. As a result, the downstream classification accuracy remains low despite the seemingly good generative metric scores.}

\rev{The quantitative analysis of the generated data provides new insights into the behavior of DiffDA methods. Due to the mismatch between existing generative scores and the ultimate metric of downstream classification accuracy, evaluating a DiffDA method still requires training with the generated data and testing the resulting classifier. If suitable quantitative metrics for evaluating generated samples in DiffDA tasks can be developed, they would substantially reduce the computational cost of training and testing downstream classifiers during method design and hyperparameter tuning. This direction could be the focus of the future work.}

\textbf{Classifier Architectures}. To assess the generality of DiffDA methods, we evaluate their performance with different classifier backbones: MobileNetV3-Large (5.4M parameters), ResNet-50 (25.6M), and ViT-B/16 (86M). Experiments are conducted on two settings: Caltech-101, where classifiers are trained from scratch, and the 5-shot classification task of the Birds dataset, where classifiers are initialized with ImageNet-pre-trained weights. The results, shown in Table~\ref{tab:backbone}, demonstrate that DiffDA consistently enhances performance across all architectures. Notably, smaller models show greater relative improvement, highlighting the potential of DiffDA for improving performance in compute-limited settings such as edge deployment.

\begin{center}
  \captionof{table}{Top-1 accuracy (\%) of DiffDA methods using MobileNetV3-Large (MN), ResNet-50 (RN), and ViT-B/16 (ViT) on \textbf{Caltech-101} and on \textbf{Birds(5-Shot)}.}
  \label{tab:backbone}

  \begin{minipage}[t]{0.45\textwidth}
    \centering
    \tablebodyfont
    {\setlength{\tabcolsep}{1.5pt}
    \begin{tabular}{cccc}
      \toprule
      \textbf{Caltech-101} & \textbf{MN} & \textbf{RN} & \textbf{ViT} \\
      \midrule
      Baseline & 60.11 & 54.33 & 26.29 \\
      \midrule
      \multirow{2}{*}{Real Guidance} & 72.35 & 65.38 & 33.71 \\
      & (\up{\textbf{+12.24}}) & (\up{+11.05}) & (\up{+7.42}) \\
      \midrule
      \multirow{2}{*}{GIF} & 75.18 & 76.49 & 39.77 \\
      & (\up{+14.07}) & (\up{\textbf{+22.16}}) & (\up{+13.48}) \\
      \midrule
      \multirow{2}{*}{DiffMix} & 77.00 & 77.62 & 40.51 \\
      & (\up{+16.89}) & (\up{\textbf{+23.29}}) & (\up{+14.22}) \\
      \bottomrule
    \end{tabular}
    }
  \end{minipage}%
  \hspace{0.05 \textwidth}
  \begin{minipage}[t]{0.45\textwidth}
    \centering
    \tablebodyfont
    {\setlength{\tabcolsep}{1.5pt}
    \begin{tabular}{cccc}
      \toprule
      \textbf{Birds(5-Shot)} & \textbf{MN} & \textbf{RN} & \textbf{ViT} \\
      \midrule
      Baseline & 33.26 & 52.50 & 58.09 \\
      \midrule
      \multirow{2}{*}{DiffuseMix} & 36.42 & 55.50 & 58.47 \\
       & (\up{\textbf{+3.16}}) & (\up{+3.00}) & (\up{+0.38}) \\
      \midrule
      \multirow{2}{*}{DA-Fusion} & 39.32 & 56.04 & 59.84 \\
       & (\up{\textbf{+3.06}}) & (\up{+2.54}) & (\up{+1.75}) \\
      \midrule
      \multirow{2}{*}{Diff-Mix} & 44.10 & 62.91 & 66.09 \\
       & (\up{\textbf{+10.84}}) & (\up{+10.41}) & (\up{+8.00}) \\
      \bottomrule
    \end{tabular}
    }
  \end{minipage}%
\end{center}

\textbf{Computational Cost}. All experiments are conducted on NVIDIA RTX 4090D GPUs. Diffusion model fine-tuning is performed with two GPUs in parallel, while sample generation and classifier training are executed on a single GPU. All fine-tuning follows the same training recipe, resulting in consistent time costs across datasets. In contrast, the time required for sample generation and classifier training depends on the size of the original dataset. \rev{Since we adopt the same sample utilization strategy for all DiffDA methods, the time spent on classifier training is identical across methods for a given dataset.} We report the GPU hours consumed by each DiffDA method on the Caltech-101, CIFAR-100, and ImageNet-100 datasets, which contain 2,425, 10,000, and 25,000 real samples respectively. The reported time includes all three stages: model fine-tuning, sample generation, and classifier training. Detailed results are presented in Table~\ref{tab:time}. The computational cost of DiffDA methods primarily stems from the sample generation stage, but for larger-scale datasets, the choice of sample utilization strategy also has a notable impact.

\vspace{-0.6cm}
\begin{table}[h]
\centering
\caption{GPU hours required by each DiffDA method on Caltech-101 (CA), CIFAR-100 (CI), and ImageNet-100 (IN) datasets, broken down into fine-tuning (FT), generation (GEN), and classification (CLS) stages.}
\label{tab:time}
{\setlength{\tabcolsep}{5pt}
\begin{tabular}{c|c|ccc|ccc|ccc}
\toprule
\multirow{2}{*}{\textbf{Method}} & \multirow{2}{*}{\textbf{FT}} & \multicolumn{3}{c|}{\textbf{GEN}} & \multicolumn{3}{c|}{\textbf{CLS}} & \multicolumn{3}{c}{\textbf{Total}} \\
\cmidrule(lr){3-5} \cmidrule(lr){6-8} \cmidrule(lr){9-11}
 & & \textbf{CA} & \textbf{CI} & \textbf{IN} & \textbf{CA} & \textbf{CI} & \textbf{IN} & \textbf{CA} & \textbf{CI} & \textbf{IN} \\
\midrule
Baseline      & 0.00 & 0.00 & 0.00 & 0.00 & 0.18 & 0.54 & 1.02 & 0.18 & 0.54 & 1.02 \\
Real Guidance & 0.00 & 1.91 & 7.70 & 20.07 & 0.68 & 2.26 & 5.11 & 2.59 & 9.96 & 25.18 \\
GIF           & 0.00 & 2.26 & 9.21 & 22.59 & 0.68 & 2.26 & 5.11 & 2.94 & 11.47 & 27.70 \\
DiffuseMix    & 0.00 & 2.56 & 10.53 & 26.28 & 0.68 & 2.26 & 5.11 & 3.24 & 12.79 & 31.39 \\
DA-Fusion     & 1.98 & 1.92 & 7.71 & 20.17 & 0.68 & 2.26 & 5.11 & 4.58 & 11.95 & 27.26 \\
Diff-Aug      & 3.26 & 1.94 & 7.73 & 20.19 & 0.68 & 2.26 & 5.11 & 5.88 & 13.25 & 28.56 \\
Diff-Mix      & 3.26 & 1.95 & 7.72 & 20.23 & 0.68 & 2.26 & 5.11 & 5.89 & 13.24 & 28.60 \\
\bottomrule
\end{tabular}
}
\end{table}
\vspace{-0.9cm}

\subsection{Methodological Explorations}

\rev{The proposed framework not only enables a unified analysis of existing approaches but also provides a basis for exploring improved DiffDA designs under diverse settings. In this section, we present new methodological explorations along all three components of the framework: (1) prompt engineering for both model fine-tuning and sample generation, (2) acceleration of the generation process through few-step inference, and (3) post-generation sample filtering strategies in the sample utilization stage.}

\rev{\textbf{Prompt engineering.} In the text-guided image-to-image transition pipeline of DiffDA, the choice of prompts has a significant impact on the final performance. We first examine the prompts used during diffusion model fine-tuning. In our experiments, the default prompts take the simple form ``\texttt{a photo of a <class-name>}'', which can be directly generated from the class labels of the training images. We then explore whether the fine-tuning process can benefit from more detailed textual descriptions. To this end, we employ Qwen3-VL to annotate the training images with additional textual information, resulting in prompts of the form ``\texttt{a photo of a <class-name>, <suffix>}'', where \texttt{<suffix>} provides supplementary information beyond the class label, such as the overall style of the image or the environment in which the object appears. We replace the generative backbones in Diff-Aug and Diff-Mix with models fine-tuned using these enriched prompts, a trial we refer to as \textit{suffix-enrich}, and evaluate their performance on multiple tasks. However, as shown in Table~\ref{tab:prompt}, the performance decreases, particularly on fine-grained classification tasks. We suspect that the more complex annotations may hinder the diffusion model from effectively learning new concepts. This finding suggests that simpler and more focused prompts are more suitable for DiffDA fine-tuning.}

\rev{In the sample generation stage, we similarly extend the simple prompt with \texttt{<suffix>}. We consider two strategies for its selection. The first strategy uses ChatGPT to generate descriptive phrases from imagination, allowing it to infer what kinds of descriptions might be appropriate for images of each class. The second strategy performs suffix exchange within the training set, given that we already have annotations for every sample. From the results in Table~\ref{tab:prompt}, we find that both strategies contribute to performance improvement. The former, referred to as \textit{suffix-dream}, performs better on coarse-grained classification tasks, whereas the latter, called \textit{suffix-exchange}, yields better results on fine-grained tasks. This difference is related to the characteristics of the datasets, since fine-grained datasets typically contain images with more homogeneous backgrounds and styles, and the LLM-generated descriptions may deviate from the original data distribution.}

\begin{table}[t]
\centering
\caption{Top-1 accuracy (\%) of different prompt engineering trials.}
{\setlength{\tabcolsep}{2.5pt}
\begin{tabular}{lccccc}
\toprule
\textbf{Method} & \textbf{Trial} & \textbf{Caltech-101} & \textbf{CIFAR-100} & \textbf{Birds (5-shot)} & \textbf{Birds (10-shot)} \\
\midrule
Diff-Aug & base & 71.67 & 67.73 & 60.36 & 71.71 \\
Diff-Aug & suffix-enrich & 70.24 & 66.91 & 54.87 & 66.95 \\
Diff-Aug & suffix-dream & \textbf{73.03} & \textbf{69.62} & 60.78 & 71.82 \\
Diff-Aug & suffix-exchange & 72.16 & 69.15 & \textbf{61.19} & \textbf{72.13} \\
\midrule
Diff-Mix & base & 75.62 & 71.45 & 62.96 & 73.26 \\
Diff-Mix & suffix-enrich & 72.54 & 69.13 & 56.29 & 68.81 \\
Diff-Mix & suffix-dream & \textbf{77.23} & \textbf{73.01} & 63.14 & 73.61 \\
Diff-Mix & suffix-exchange & 76.83 & 72.89 & \textbf{63.65} & \textbf{74.04} \\
\bottomrule
\end{tabular}
}
\vspace{-0.5cm}
\label{tab:prompt}
\end{table}

\rev{\textbf{Accelerating generation.} We observe that a major limitation preventing existing DiffDA methods from being applied to larger-scale datasets lies in the time cost of the generation process. On an NVIDIA RTX 4090D GPU, augmenting each set of 10,000 real images by five times requires approximately 8 GPU hours. For mainstream DiffDA methods based on the SDEdit paradigm, the computational cost of generation depends on the image-to-image transition strength $s$ and the total number of diffusion steps $T$, since $sT$ represents the number of actual noise addition and denoising steps performed during generation. Because the choice of $s$ is highly critical (as discussed in our analysis of this hyperparameter) and cannot be easily changed, we explore whether reducing $T$ can accelerate the generation process.} 

\begin{figure}[h]
    \centering
    \includegraphics[width=0.9\linewidth]{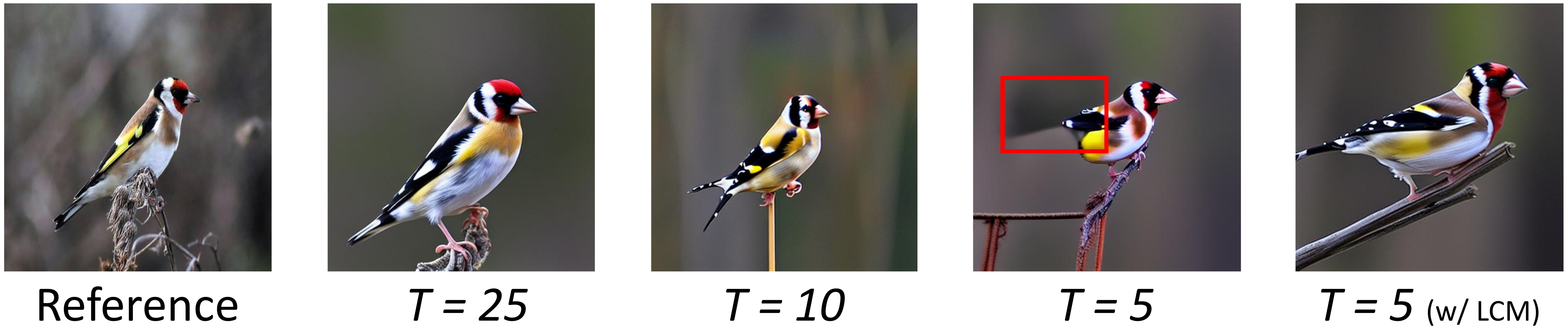}
    \caption{Examples of ``European Goldfinch'' images from the Birds dataset generated by the Diff-Aug method with different numbers of diffusion steps $T$.}
    \label{fig:few-step}
    \vspace{-0.5cm}
\end{figure}

\rev{The results shown in Table~\ref{tab:few-step} are very promising. For representative DiffDA methods, reducing $T$ from the original 25 to 10 has almost no impact on the final classification performance while providing nearly a 2.5$\times$ speedup. With the help of Latent Consistency Models (LCMs) \citep{luo2023latent}, which are fine-tuned on pretrained diffusion models to enable few-step generation, we can further reduce $T$ to 5, achieving approximately a 5$\times$ acceleration with only a small and acceptable drop in performance. This provides a practical trade-off for applying DiffDA methods in real-world scenarios. We provide examples of the generated images in Fig.~\ref{fig:few-step} to illustrate this phenomenon. Although using a larger $T$ improves the visual quality of the generated images by reducing artifacts, the training of downstream classifiers does not strictly depend on high visual quality. It only requires that the generated images contain semantically relevant information.}

\begin{center}
  \captionof{table}{Top-1 accuracy (\%) of DiffDA methods with different diffusion steps $T$.}
  \label{tab:few-step}
  \begin{minipage}[t]{0.45\textwidth}
    \centering
    \tablebodyfont
    {\setlength{\tabcolsep}{2.5pt}
    \begin{tabular}{cccc}
    \toprule
    {\textbf{Methods}} & {\textbf{$T$}} & {\textbf{Caltech}} & {\textbf{CIFAR}}  \\
    \midrule
    \multirow{3}{*}{Real Guidance} & \underline{25} & 65.38 & 66.24  \\
         &  10 & 65.26 & 66.57 \\
         &  5 & 64.25 & 66.00 \\
    \midrule
    \multirow{2}{*}{Real Guidance*}     & 5 & 64.78 & 66.37 \\
         & 3 & 63.57 & 65.33 \\
    \midrule
    \multirow{3}{*}{Diff-Aug} & \underline{25} & 71.67 & 67.73  \\
         &  10 & 71.42 & 67.69 \\
         &  5 & 69.75 & 67.47 \\
    \midrule
    \multirow{2}{*}{Diff-Aug*}     & 5 & 70.86 & 67.83 \\
         & 3 & 67.71 & 66.56 \\
    \bottomrule
    \end{tabular}
    }
  \end{minipage}%
  \hspace{-0.01 \textwidth}
  \begin{minipage}[t]{0.45\textwidth}
    \centering
    \tablebodyfont
    {\setlength{\tabcolsep}{1.5pt}
    \begin{tabular}{cccc}
    \toprule
    {\textbf{Methods}} & {$T$} & {\textbf{Birds (5-shot)}} & {\textbf{Birds (10-shot)}}  \\
    \midrule
    \multirow{3}{*}{Diff-Aug} & \underline{25} & 60.36 & 71.71  \\
         &  10 & 59.93 & 71.28 \\
         &  5 & 57.64 & 69.83 \\
    \midrule
    \multirow{2}{*}{Diff-Aug*}     & 5 & 59.46 & 70.83 \\
         & 3 & 56.71 & 68.33 \\
    \midrule
    \multirow{3}{*}{Diff-Mix} & \underline{25} & 62.96 & 73.26  \\
         &  10 & 62.77 & 72.82 \\
         &  5 & 58.85 & 71.07 \\
    \midrule
    \multirow{2}{*}{Diff-Mix*}     & 5 & 62.04 & 72.26 \\
         & 3 & 59.30 & 71.14 \\
    \bottomrule
    \end{tabular}
    }
  \end{minipage}%
  \vspace{2pt}
  \parbox{0.75\textwidth}{
\tablebodyfont * indicates the use of Latent Consistency Models (LCMs). 
}
\vspace{-0.5cm}
\end{center}

\rev{\textbf{Sample filtering.} Due to the inherent randomness of the generation process, the produced samples inevitably contain some badly generated images, especially when a high transition strength is used to encourage greater diversity. This naturally raises the question of whether filtering can be introduced during the sample utilization stage to eliminate such low-quality samples. We conduct a detailed analysis of this idea.}

\rev{The framework we study for filtering is as follows. Given the generated images and their corresponding labels, we introduce a scoring model that assigns a score to each generated sample, after which a specified proportion of the lowest-scoring samples are removed. We experiment with two types of scoring models: a base model trained only on real samples and a CLIP model. For the base model, we use its prediction on each generated image and take the probability of the target class as the score, which we refer to as \textit{Base-Prob}. For the CLIP model, we explore two scoring strategies. The first, termed \textit{CLIP-Multi}, constructs text prompts for all classes in the form of “\texttt{A photo of a <class-name>}” and matches them with the images, using the normalized score of the target class. The second, called \textit{CLIP-Binary}, forms positive and negative pairs in the form of “\texttt{A photo of a <target-class>}” and “\texttt{A photo with no <target-class>}”. For fine-grained classification tasks, we modify the positive and negative prompts to “\texttt{A photo of a <target-class>, a type of <meta-class>}” and “\texttt{A photo with no <meta-class>}”, respectively.}

\begin{figure}[h]
    \centering
    \includegraphics[width=0.99\linewidth]{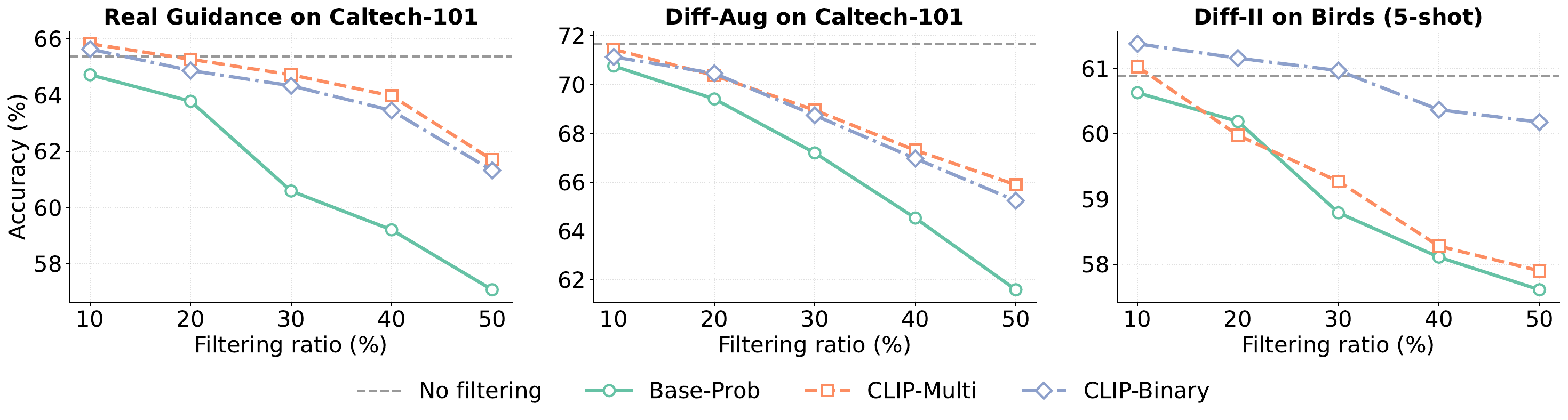}
    \caption{Comparison of filtering strategies under various filtering ratios.}
    \label{fig:filtering}
     \vspace{-0.5cm}
\end{figure}

\rev{Figure~\ref{fig:filtering} presents the results of different filtering strategies under various filtering ratios. The results suggest that sample filtering is not always an effective post-processing strategy. For coarse-grained conventional classification tasks, where all generated images are added to the training set and classifiers are trained from scratch, the benefit of removing potentially low-quality samples is often outweighed by the loss caused by reducing the overall training data size. This trend becomes more pronounced as the filtering ratio increases. For fine-grained few-shot classification tasks, although we maintain a constant training data size through random replacement, filtering can still mistakenly discard some informative generated samples. These observations are consistent with the findings of \citet{he2022realguidance}, which emphasize that instead of filtering samples after generation, incorporating more appropriate guidance during the generation process is a more effective way to produce useful training data.}

\rev{Nevertheless, we can still compare the relative effectiveness of different filtering strategies. Using the base model itself as the scoring model proves to be suboptimal. In data-scarce scenarios, the base model tends to overfit the limited real samples, and relying on its predicted probabilities for filtering often leads to the mistaken removal of hard samples that differ from the real data but are still beneficial for improving classification performance. The CLIP model performs relatively better on coarse-grained conventional classification tasks, as it possesses strong recognition ability for common categories and provides a complementary perspective to the base model. However, on fine-grained tasks, CLIP is also constrained by its lack of detailed concept knowledge. In such cases, the modified \textit{CLIP-Binary} strategy exhibits a clear advantage, since it does not need to identify each specific bird species but only needs to filter out images that fail to successfully generate birds.}

\rev{
The above experiments and analyses collectively demonstrate the novelty of our work in exploring general improvement techniques for DiffDA. By combining these methodological enhancements, we are able to develop more efficient and more effective DiffDA methods tailored to different task requirements. For example, on the ImageNet-100 dataset, applying the \textit{suffix-dream} prompts to the Diff-Mix method, along with a reduced number of diffusion steps ($T = 5$ with LCM), decreases the generation time from approximately 20 GPU hours to about 4 GPU hours, while simultaneously improving classification accuracy from 78.29\% to 79.87\%. On the Aircraft 10-shot task, Diff-Mix with the \textit{suffix-exchange} prompts, combined with the \textit{CLIP-Binary} filtering strategy that removes 10\% of generated samples, further improves the classification accuracy from 59.37\% to 60.79\%.
}

\section{Conclusion}\label{sec5:conclusion}
In this work, we introduce UniDiffDA, a unified analytical framework that decomposes diffusion-based data augmentation into three core components: model fine-tuning, sample generation, and sample utilization. Using this framework, we establish a comprehensive and fair evaluation protocol and benchmark representative DiffDA methods across diverse low-data classification tasks, including generic, fine-grained, long-tailed, multi-domain, and medical settings. Our study shows that there is no universally best method and that DiffDA effectiveness depends on the interaction between its core components and the characteristics of the target task. \rev{Building on the proposed framework, we further explore general techniques that make existing DiffDA methods more effective and more efficient.} We release all implementations and configurations in a unified, reproducible codebase to support future research and practical deployment of improved DiffDA techniques.

\backmatter








\section*{Declarations}

\textbf{Funding}. This work was supported by National Science and Technology Major Project (2023ZD0120700) and NSFC Project (62222604).

\noindent \textbf{Competing interests}. The authors have no competing interests to declare that are relevant to the content of this article.

\noindent \textbf{Data availability}. All datasets used in this study are publicly available. The specific data splits used in our experiments have been released at \url{https://huggingface.co/datasets/nukezil/DiffDA-Eval}.

\noindent \textbf{Code availability}. The code used in this study is available at \url{https://github.com/nukezil/DiffDA-Eval}.

\noindent \textbf{Author contribution}. Zekun Li conducted the core research, including experiment design, implementation, and writing. Yinghuan Shi supervised the study and provided critical feedback. Yang Gao offered overall guidance and project oversight. Dong Xu provided critical feedback and revised the manuscript.

\noindent \textbf{Ethics approval and consent to participate}. Not applicable.

\noindent \textbf{Consent for publication}. Not applicable.

\noindent \textbf{Materials availability}. Not applicable.


\noindent

\bigskip

\bibliography{sn-bibliography}

\end{document}